\documentclass[10pt]{article}
\usepackage{epsfig,amsmath,amsthm,graphicx,amssymb,overpic,cite}
\usepackage{listings,diagbox,xcolor,empheq}

\def\be{\begin{equation}}
\def\ee{\end{equation}}
\def\bee{\begin{eqnarray}}
\def\ene{\end{eqnarray}}
\def\bes{\begin{subequations}}
\def\ees{\end{subequations}}

\def\no{\nonumber}

\def\v{\vspace{0.1in}}
\def\d{\displaystyle}

\begin{document}

%%%%%%%%%%%%%%%%%%%%%%%%%%%%%%%%%%%%%%%%%%%%%%%%%%%%%%%%%%%%%%%%%%%%%%%%%%%%%%%%%
\baselineskip=13pt
\renewcommand {\thefootnote}{\dag}
\renewcommand {\thefootnote}{\ddag}
\renewcommand {\thefootnote}{ }

\pagestyle{plain}

\begin{center}
\baselineskip=16pt \leftline{} \vspace{-.3in} {\Large \bf
Data-driven discoveries of B\"acklund transforms and soliton evolution equations via deep neural networks learning} \\[0.2in]
\end{center}

\begin{center}
Zijian Zhou$^{1,2}$, Li Wang$^{3,4}$, Weifang Weng$^{1,2}$, and Zhenya Yan$^{1,2,*}$\footnote{$^{*}${\it Email address}: zyyan@mmrc.iss.ac.cn (Corresponding author)}  \\[0.05in]
{\it \small $^{1}$Key Laboratory of Mathematics Mechanization, Academy of Mathematics and Systems Science, \\ Chinese Academy of Sciences, Beijing 100190, China \\
 $^{2}$School of Mathematical Sciences, University of Chinese Academy of Sciences, Beijing 100049, China\\
 $^{3}$Yanqi Lake Beijing Institute of Mathematical Sciences and Applications, Beijing, 101408, China \\
  $^{4}$Yau Mathematical Sciences Center and Department of Mathematics, Tsinghua University, Beijing, 100084, China} \\
 %(Date:\,\, \today)
\end{center}

%\vspace{0.1in}

{\baselineskip=13pt

%\begin{tabular}{p{16cm}}
% \hline \\
%\end{tabular}

\vspace{0.18in}

%\begin{abstract} \small \baselineskip=12pt
\noindent {\bf Abstract}\, We introduce a deep neural network learning scheme to discover the B\"acklund transforms (BTs) of soliton evolution equations and an enhanced deep learning scheme for data-driven soliton equation discovery, respectively. The first deep learning scheme takes advantage of some solution (or soliton equation) informations to train the data-driven BT discovery, and is valid in the study of the BT of the sine-Gordon equation, and complex and real Miura transforms between the defocusing (focusing) mKdV equation and KdV equation, as well as the data-driven mKdV equation discovery via the Miura transforms. The second deep learning scheme uses the higher-order solitons generated by the explicit/implicit BTs to study the data-driven discoveries of mKdV and sine-Gordon equations, in which the
high-order soliton informations are more powerful for the enhanced leaning soliton equations with higher accuracies.

%\vspace{0.1in} \noindent MSC: 37K15; 35Q53; 35Q15; 37K40

\vspace{0.1in} %\noindent

%\end{*abstract}

\vspace{0.18in}

%\vspace{-0.05in}
%\begin{tabular}{p{16cm}}
%  \hline \\
%\end{tabular}
\baselineskip=13pt

\section{Introduction}

In the fields of applied mathematics and nonlinear mathematical physics, there are many types of physically interesting nonlinear evolution partial differential equations (PDEs). Particularly, since the well-known  Korteweg-de Vries (KdV) equation with solitary waves was presented by Korteweg and de Vries~\cite{kdv}, various of soliton evolution equations (e.g., the Boussinesq equation, mKdV equation, KP equation, sine-Gordon equation, nonlinear Schr\"odinger equation, Gross-Pitaevskii equation)~\cite{soliton} play the important roles in the fields of nonlinear science, such as fluid mechanics, nonlinear optics, quantum optics, Bose-Einstein condensates, plasmas physics, ocean, atmosphere,
biology, and even finance~\cite{hase,nailbook,agrawal,bec,kharif,osborne,yanfrw}. Many types of analytical, numerical and experimental approaches have been used to deeply explore the wave structures and properties of these soliton equations (see, e.g., Refs.~\cite{soliton,hase,nailbook,agrawal,bec,kharif,osborne,yangbook} and references therein).

Since B\"acklund~\cite{back1} first found a transform (alias the auto-B\"acklund transform (aBT)) of the sine-Gordon equation $u_{xt}=\sin u$ in 1875, and Darboux~\cite{DT82} found a transform (alias Darboux transform) of the Strum-Liouville equation (alias the linear Schr\"odinger equation) $\psi_{xx}+[\lambda-V(x)]\psi=0$ in 1882,
many types of well-known analytical transforms were found between the same equation or different equations~\cite{BT76,lamb74,BT-book2002,DT1991, bluman1989,Hirota-book}. For example, Hopf~\cite{Hopf} and Cole~\cite{Cole} independently established a BT between the nonlinear Burgers equation $u_t+uu_x-\mu u_{xx}$ and linear heat (or diffusion) equation $v_t-\mu v_{xx}=0$ in 1950-1951.  In 1967,  Gardner, Greene, Kruskal, and Miura (GGKM)~\cite{Gardner1967} solved the initial value problem of the KdV equation starting from its coupled linear PDEs, which are just its Lax pair~\cite{soliton}. In 1968, strongly motivated by the GGKM's idea,  Lax~\cite{Lax} presented a general formal BT (alias Lax pair) implying that the eigenvalues of the linear operator $\{L(\phi)\psi=\lambda\psi(\phi),\, \psi_t=B(\phi)\psi\}$ are integrals of the nonlinear equation $\phi_t=K(\phi)$. In the same year, Miura~\cite{miura} found a new BT (alias the Miura transform (MT)) between the KdV equation
$v_t+6vv_x+v_{xxx}=0$ and focusing (or defocusing) mKdV equation $u_t\pm 6u^2u_x+u_{xxx}=0$.
%In 1971, Hirota~\cite{hirota71} found a new BT (alias Hirota bilinear transform) changing the KdV equation into a bilinear equation.

With the quick development of cloud computing resources and mass data, deep learning~\cite{DL1,DL2} has been used in
many fields containing cognitive science~\cite{DL3}, image recognition~\cite{DL4}, genomics~\cite{DL5},
industrial areas~\cite{DL6, DL7}, and etc. In particular, in the past of decades, some deep neural network learning methods~\cite{dl-pde1,dl-pde2,dl-pde3,dl-pde4,dl-pde5,dl-pde6,Han,raiss18,raiss19} have been
developed to study the partial differential equations (PDEs), which play an important role in the various of scientific fields.
The powerful physics-informed neural network (PINN) method~\cite{raiss18,raiss19} was used to investigate the  PDEs~\cite{raiss20,pinn1,pde20,yan-pla21a,yan-pla21b,wang-peakon21,zhou-rw21,chen2021,libiao2022}, the fractional PDEs~\cite{fpde}, and stochastic PDEs~\cite{spde}.

In this paper, we would like to develop two kinds of deep neural network learning methods to study the data-driven discoveries of BTs and soliton equations via the general system
\begin{subequations}\label{sys}
\begin{empheq}[left=\empheqlbrace]{align}
 &F(u,u_t,u_x,u_{tt},u_{tx},u_{xx},\cdots)=0, \label{PDE1} \v\\
&G(u',u'_t,u'_x,u'_{tt},u'_{tx},u'_{xx},\cdots)=0, \label{PDE2} \v\\
&\phi_i(u,u_t,u_x,u_{tt},u_{tx},u_{xx},\cdots,u',u'_t,u'_x,u'_{tt},u'_{tx},u'_{xx},\cdots)=0, \quad (i=1,2,...), \label{bkt}
\end{empheq}
\end{subequations}
where $u=u(x,t),\, u'=u'(x,t)$, the considered spario-temporal region is $(x,t)\in [-L, L]\times [-T, T]$, Eq.~(\ref{bkt}) is called the BT between Eqs.~(\ref{PDE1}) and (\ref{PDE2}). In particular, if Eq.~(\ref{PDE2}) is equivalent to Eq.~(\ref{PDE1}), then the transform (\ref{bkt}) is called the aBT. It is obvious to see that the information of Eq.~(\ref{PDE1}) can be shifted to Eq.~(\ref{PDE2}) with the aid of BT (\ref{bkt}). Sometimes, the structure of the transformed Eq.~(\ref{PDE2}) may become simpler. Therefore, BTs can be used to discover the new informations between Eqs.~(\ref{PDE1}) and (\ref{PDE2}).

The rest of this paper is organized as follows. In Sec. 2, we will introduce a deep neural network learning scheme to study aBTs and BTs, e.g., the aBT of the sine-Gordon equation, and Miura transform between the focusing/defocusing mKdV equation and KdV equation. Moreover, the deep learning scheme can also be used to discover soliton equations with the aid of BTs.
In Sec. 3, a new deep learning method discovering the soliton equations is displayed based on the BTs. We use the implicit
and explicit BTs to exhibit the data-driven discoveries of the sine-Gordon equation and mKdV equation with the aid of the aBT of the sine-Gordon equation, and Darboux transform of the focusing mKdV equation, respectively. Finally, some conclusions and discussions are summarized in Sec. 4.

\section{Data-driven discoveries of BTs and soliton equations}

\subsection{Deep learning scheme discovering the BTs and equations via BTs }}

We here would like to introduce the deep learning scheme for the discoveries of BTs and soliton equations by examining system (\ref{sys}). The main idea of this scheme is to
use some constraints on $u(x,t)$ and $u'(x,t)$ to find the approximate transform between $u(x,t)$
and $u'(x,t)$, where $u(x,t)$ and $u'(x,t)$ are represented by one or two deep neural networks. There are many kinds of neural networks including the fully-connected neural network, convolution neural network, and recurrent neural network.
The constraints of the functions include the real solution data-set and the corresponding equations.
The aim of the scheme is to make the neural network solution approach to the real data better and match some physical laws efficiently.

\begin{figure}[!t]
%\hspace{2.5in}
\begin{center}
 \hspace{0.35in} {\scalebox{0.75}[0.75]{\includegraphics{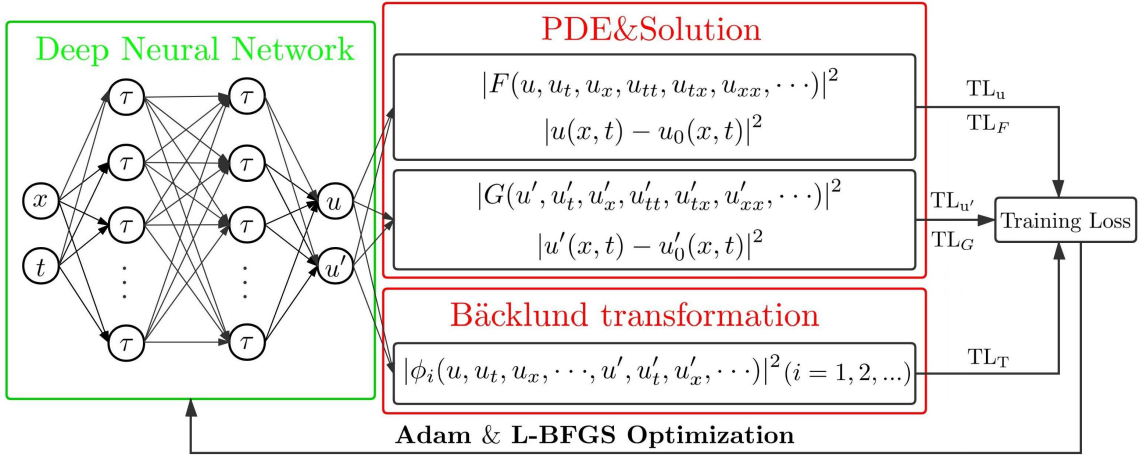}}}
 \end{center}
\par
\vspace{-0.05in}
\caption{\small The deep learning scheme for the data-driven discovery of BTs and soliton equations.}
\label{fig1-DNN}
\end{figure}

Figure~\ref{fig1-DNN} displays the deep learning scheme of the BT discovery, where $u(x,t)$ and $u'(x,t)$ are represented by
a deep neural network, which is, in general, chosen as a fully-connected neural network. It will share the same network parameters, weights, and biases. For some conditions, they can be represented by two different networks to eliminate mutual influence.
In this diagram, we assume that $u(x,t)$ and $u'(x,t)$ are real-valued functions. If the solutions of Eq.~(\ref{PDE1})
are complex, the number of output neurons would be double. The number of input neurons equals to the independent variables of $u(x,t)$ and $u'(x,t)$. $\tau$ represents the activation function, and is chosen as $\tau(x)=\tanh(x)$ in this scheme, whose aim is
to add the nonlinear action to the deep neural network. Notice that one can also choose other types of activation functions, such as the sigmoid (logistic) function, threshold function, piecewise linear function, ReLU function, ELU function, swish function, and softmax function~\cite{wang-peakon21}.

The loss function during the training process consists of three different parts. It can be simply written as:
\bee\label{TL}
{\rm TL}={\rm TL}_T+{\rm TL}_{F/u}+{\rm TL}_{G/u'},
\ene
which can be called T-part (${\rm TL}_T$), F-part (${\rm TL}_{F/u}$), and G-part (${\rm TL}_{G/u'}$), and are defined by
\begin{equation} \label{TLu}
\begin{aligned}
    {\rm TL}_F&=\frac{1}{N_p}\sum_{j=1}^{N_p}|F(u(x_j,t_j),u_t(x_j,t_j),u_x(x_j,t_j),\cdots)|^2,\\
    \quad {\rm TL}_u&=\frac{1}{N_p}\sum_{j=1}^{N_p}|u(x_j,t_j)-u_0(x_j,t_j)|^2,\\
    {\rm TL}_G&=\frac{1}{N_p}\sum_{j=1}^{N_p}|G(u'(x_j,t_j),u'_t(x_j,t_j),u'_x(x_j,t_j),\cdots)|^2,\\
    \quad {\rm TL}_{u'}&=\frac{1}{N_p}\sum_{j=1}^{N_p}|u'(x_j,t_j)-u'_0(x_j,t_j)|^2,\\
    {\rm TL}_{T}&=\frac{1}{N_p}\sum_{j=1}^{N_p}|\phi_j(u,u_t,u_x,\cdots, u',u'_t,u'_x,\cdots)|^2_{(x,t)=(x_j,t_j)},
    \end{aligned}
\end{equation}
where $\{x_j,\, t_j,\}_{j=1}^{N_p}$ denote the set of sampling points in some spatio-temporal region $(x,t)\in [-L, L]\times [-T, T]$, $N_p$ stands for the number of sampling points, which are generated by using the Latin Hypercube Sampling strategy~\cite{stein87}, and $\{u_0(x_j,t_j)\}_{j=1}^{N_p}$ and $\{u'_0(x_j,t_j)\}_{j=1}^{N_p}$ represent the sampling solution data of Eqs.~(\ref{PDE1}) and (\ref{PDE2}), respectively.

The data type determines the selection of the loss function. To make the neural network satisfy the structures of the equations and BTs, the loss function should contain the above three parts. Specifically, if we assume that the discovery problem of BTs contains both data sets, $\{u_0(x_j,t_j)\}_{j=1}^{N_p}$ and $\{u'_0(x_j,t_j)\}_{j=1}^{N_p}$, then all terms of the loss function (\ref{TLu}) can be add to the final loss function (\ref{TL}). This case is displayed in the subsection \ref{learnmu}. But if it only contains one data-set, the algorithm also works when the equation loss (${\rm TL}_{F}$ or ${\rm TL}_{G}$) of missing data-set funtion is added to the final loss function. For instance, the only data-set $\{u_0(x_j,t_j)\}_{j=1}^{N_p}$ is known in the following subsection \ref{learnbt}  about the BT discovery of the sine-Gordon equation.

Based on this framework, we can also study another question: learning the equation from the data-set of another equation by a BT. Generally speaking, if the BT is explicit, then the given data-set can be easily transformed into the data-set of another unknown function. But if the BT is implicit, it is not easy to study this question (without loss of generality, we assume that $u'_0(x,t)$ is implicit with respect to $u_0(x,t)$). In our framework, although the BT is implicit, the T-part loss function can convert the data-set of $u_0(x,t)$ to the data-set of $u'_0(x,t)$. The estimated data-set of $u'_0(x,t)$ will be represented as the neural network approximated solution, and the loss function ${\rm TL}_{G}$ will be used to train the unknown parameters in the equation $G=0$ to approach the right values. The subsection \ref{viamu} will verify the effectiveness of our framework in terms of some examples.

In the scheme, firstly, for a set of larger sampling points, we would like to use an efficient mini-batch optimization algorithm Adam~\cite{Adam}. Secondly, the model will be trained by a full-batch optimization algorithm L-BFGS~\cite{BFGS} until the difference of the loss function is less than the Machine Epsilon. In what follows, we will use some examples to verify the validity of our deep learning scheme. \v

\subsection{Examples of the data-driven BT/MT discoveries}

 In what follows, we would like to use the known equations (e.g., sine-Gordon equation (\ref{sG}), the focusing mKdV equation (\ref{mKdV}) with the KdV equation (\ref{KdV}), or defocusing mKdV equation (\ref{mKdV2}) with the KdV equation (\ref{KdV})) and their corresponding known solution (\ref{sG-solu}),  (\ref{mKdV-solu}) with (\ref{KdV-solu}), or   (\ref{mKdV-solu1}) with (\ref{KdV-solu1}) to discover
the unknown BT (\ref{sG-back1}), the complex Miura transform (\ref{miura4}), or real Miura transform  (\ref{miura5}) by using the above deep learning scheme, respectively.

\v\subsubsection{Data-driven BT discovery of the sine-Gordon equation}\label{learnbt}

The well-known sine-Gordon (s-G) equation\cite{sG}
\bee\label{sG}
 u_{xt}={\rm sin}u,\quad u(x,t)\in\mathbb{R}[x,t]
\ene
is a physically interesting model, and can be used to describe the theory of crystal dislocations, Bloch-wall motion, splay waves in lipid membranes, magnetic flux on a Josephson line, and elementary particles~\cite{sG-Rubinstein70,sG-Barone71,sG-Caudrey75}.
The s-G equation (\ref{sG}) admits the auto-B\"acklund transform (aBT)~\cite{back1}
\bee\label{sG-back}
\left\{\begin{array}{l}
 u'_x=u_x-2\beta\sin\left(\dfrac{u+u'}{2}\right), \v \\
 u'_t=-u_t+\dfrac{2}{\beta}\sin\left(\dfrac{u-u'}{2}\right),
\end{array}\right.
\ene
where $\beta\not=0$ is an arbitrary real-valued constant, that is, if $u(x,t)$ is a solution of the s-G equation (\ref{sG}), then so is $u'(x,t)$ given by Eq.~(\ref{sG-back}). In what follows, we would like to use the above-mentioned deep learning method to discover the parameters of the aBT (\ref{sG-back}). For convenience, we consider the generalized aBT
\bee\label{sG-back1}
\left\{\begin{array}{l}
 u'_x=au_x-b\sin\left(\dfrac{u+u'}{2}\right)+h uu_x, \v \\
 u'_t=-cu_t+d\sin\left(\dfrac{u-u'}{2}\right)-f uu_t,
\end{array}\right.
\ene
where the parameters $a,\, b,\, c,\, d,\, h$, and $f$ are real-valued parameters to be determined later, the two new quadratic nonlinear terms are introduced in the unknown aBT (\ref{sG-back1}).  In particular, as
$a=c=1,\, b=2\beta,\, d=2/\beta,\, h=f=0$, the unknown aBT (\ref{sG-back1}) reduces to the known exact aBT (\ref{sG-back}).

 We use the above-mentioned deep leaning scheme to discover the aBT of the s-G equation in two cases by considering the system
  \bee\label{sg-sys}
 \left\{
 \begin{array}{l}
 %u_{xt}-{\rm sin}u=0, \v\\
 u'_{xt}-{\rm sin}u'=0, \v\\
 u'_x-au_x+b\sin\left(\dfrac{u+u'}{2}\right)-h uu_x=0, \v \\
 u'_t+cu_t-d\sin\left(\dfrac{u-u'}{2}\right)+f uu_t=0
 %\v\\ u(x,0)=u_0(x),\quad  u(-L, x)=u(L, x), \v\\ %u(?L,x) = u(L,x), \v \\
% u'(x,0)=u'_0(x), \quad u'(-L, x)=u'(L, x),
 %\quad  u'(?L,x) = u'(L,x),
 \end{array} \right.
  \ene
in the spario-temporal region $(x,t)\in [-10, 10]\times [-5, 5]$, where $a,\, b,\, c,\, d,\, h$, and $f$ are parameters to be determined. The training data are generated by using the breather solution of the s-G equation (\ref{sG})~\cite{sG-Barone71}
\bee\label{sG-solu}
 u=4\arctan\left(\frac{{\rm tan}(\mu){\rm sin}(X\cos(\mu))}{{\rm cosh}(T\sin(\mu))}\right),\quad
 X=\frac{(1-k)x-(1+k)t}{\sqrt{1-k^2}},\quad T=\frac{(1-k)x+(1+k)t-x_0}{\sqrt{1-k^2}},
\ene
where $k\in [0, 1),\, \mu\not=0$. The breather solution data-set (TL$_{u}$) given by Eq.~(\ref{sG-solu}) and original equation (TL$_{f}$) given by Eq.~(\ref{sG}) together generate the F-part loss. The undetermined transforms form the T-part loss. And the G-part loss is only formed by the transformed equation (TL$_{f}$).

Here, the hidden neural network $\widehat u(x,t)$ in Python can be defined as
\begin{lstlisting}
def u(x, t):
    U = neural_net(tf.concat([x,t],1), weights, biases)
    u = U[:, 0:1]
    v = U[:, 1:2]
    return u, v
\end{lstlisting}
such that the residual neural network $f_{sG}(x,t)$ and $f_{BT}(x,t)$ in Python can be obtained as
\begin{lstlisting}
 def f_sG(x, t):
     u, v = u(x, t)
     v_t = tf.gradients(v, t)[0]
     v_x = tf.gradients(v, x)[0]
     v_xt = tf.gradients(v_x, t)[0]
     u_x = tf.gradients(u, x)[0]
     u_t = tf.gradients(u, t)[0]
     f_sG = v_xt - sin(v)
     f_BT1 = v_x - a * u_x + b * tf.sin((u + v) / 2) - h * u * u_x
     f_BT2 = v_t + c * u_t - d * tf.sin((u - v) / 2) + f * u * u_t
     return f_sG, f_BT1, f_BT2
\end{lstlisting}

\begin{figure}[!t]
\begin{center}
\vspace{0.05in} %\hspace{-0.05in}
{\scalebox{0.7}[0.65]{\includegraphics{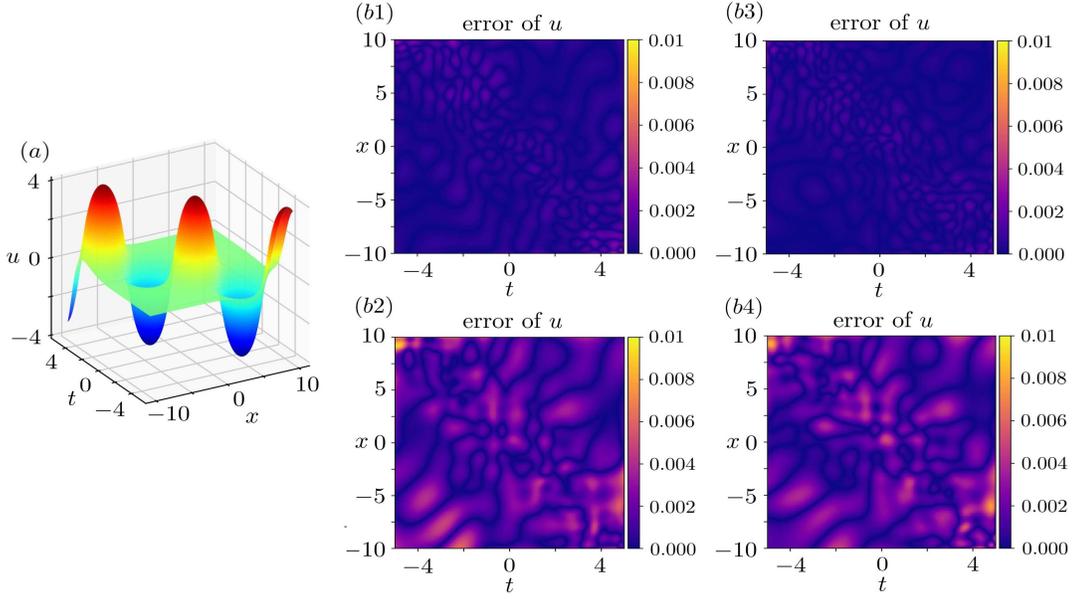}}}
\end{center}
\par
\vspace{-0.1in}
\caption{\small Data-driven aBT discovery of the sine-Gordon equation (\ref{sG}): (a) breather solution (\ref{sG-solu})
generating the training data; (b1-b4) Relative error between exact solution and neural network solution: (b1, b3)
training data without a noise; (b2, b4) training data with a $2\%$ noise.  The relative $\mathbb{L}^2-$norm errors of $u(x,t)$, respectively, are (b1) $3.5255\cdot 10^{-4}$, (b2) $1.3327\cdot 10^{-3}$, (b3) $2.7313\cdot 10^{-4}$, and (b4) $1.3390\cdot 10^{-3}$. Training times in (b1-b4) are 619.92s, 659.96s, 637.08s, and 697.76s, respectively.}
\label{fig2-SGerror}
\end{figure}

\v {\it Case A}.---In this case, we suppose that $a,\,b,\,c$, and $d$ are unknown parameters, and $h=f=0$.
 It should be pointed out that the two parameters $b$ and $d$ are not fixed. We know that $bd=4$ in the given exact aBT (\ref{sG-back}) such that we only consider the product value of $b$ and $d$ in the deep learning. We use a 6-layer neural network with 5 hidden layers and 40 neurons per layer to learn system (\ref{sg-sys}). Without loss of generality, we take the initial value of all free parameters as $1$, i.e., $a=b=c=d=1$. We choose $(x,t)\in [-10,10]\times[-5,5]$ as the training region, from which 10,000 sample points
are taken by the Latin Hypercube Sampling strategy~\cite{stein87}. Moreover, the 20,000 steps Adam and 50,000 steps L-BFGS optimizations are used in the deep learning. Fig.~\ref{fig2-SGerror}(a) displays the trained breather solution by using the deep neural network.
Case A in Table~\ref{sG-table2} exhibits the learning parameters about $a,\,c,\, b$ and $d$, and their errors under two senses of the training data without a noise and with a $2\%$ noise, respectively, which imply that the used deep learning method is effective. Moreover, the errors are exhibited in Figs.~\ref{fig2-SGerror}(b1, b2) for the cases without a noise and with a $2\%$ noise, respectively.  The training times are 619.92s and 637.08s, respectively.

\v {\it Case B}.---In this case, we suppose that $a,\,b,\,c,\, d$, $h,$ and $f$ are all unknown parameters. We used the same deep neural network method as Case A to study this case. Case B in Table~\ref{sG-table2} displays the learning parameters about $a,\,b,\,c,\, d$, $h,$ and $f$, and their errors under two senses of the training data without a noise and with a $2\%$ noise, respectively, which imply that the used deep learning method is effective. Moreover, the errors are exhibited in Figs.~\ref{fig2-SGerror}(b3, b4) for the cases without a noise and with a $2\%$ noise, respectively. The training times are 659.96s and 697.76s, respectively.

\begin{table}[!t]
	\centering
    \setlength{\tabcolsep}{10pt}% column separation
    \renewcommand{\arraystretch}{1.4}%row space
	\caption{Data-driven discovery of parameters $a$, $bd$, $c,\, h,\, f$ in aBT (\ref{sG-back1}) and errors, as well as training times. \vspace{0.25in}}
	\begin{tabular}{cccccccc} \hline\hline
	Case & $a$ & $b\cdot d$ & $c$  & $h$ & $f$ & \\  \hline
  Exact & 1 & 4 & 1& 0 & 0 & \\
  A (no noise) & 1.00007 & 4.00006  & 1.00001 & 0 & 0 & \\
  A ($2\%$ noise) & 0.99995  & 4.00057  & 1.00003 & 0 & 0 & \\
   B (no noise) & 1.00004 & 4.00021 & 1.00001 & -4.24$\times10^{-7}$ & -2.51$\times10^{-6}$ & \\
   B (2$\%$ noise) & 0.99988 & 3.99959 & 0.99985 & -3.90$\times10^{-4}$ & -3.60$\times10^{-4}$ &\\
    \hline\hline
    Case & error of $a$ & error of $b\cdot d$& error of $c$ & error of $h$ & error of $f$ & time \\  \hline
  A (no noise) & 7.44$\times10^{-5}$ & 6.45$\times10^{-5}$  & 1.31$\times10^{-5}$ & 0 & 0 &   619.92s \\
  A ($2\%$ noise) & 5.29$\times10^{-4}$  & 5.74$\times10^{-4}$ & 3.24$\times10^{-5}$ & 0 & 0 &  637.08s  \\
  B (no noise) & 4.17$\times10^{-5}$ & 2.12$\times10^{-4}$ & 1.37$\times10^{-5}$ & 4.24$\times10^{-7}$ & 2.51$\times10^{-6}$ &  659.96s  \\
  B (2$\%$ noise) & 1.20$\times10^{-4}$ & 4.05$\times10^{-4}$ & 1.47$\times10^{-4}$ & 3.94$\times10^{-4}$ & 3.57$\times10^{-4}$
  & 697.76s \\
    \hline\hline
	\end{tabular}
	\label{sG-table2}
\end{table}

\subsubsection{Data-driven discovery of Miura transforms}\label{learnmu}

In 1968, Miura~\cite{miura} presented the well-known complex Miura transform (a special BT)
\bee\label{miura}
 v=iu_x+u^2,\quad i=\sqrt{-1},
\ene
and real Miura transform
\bee\label{miura2}
 v=u_x-u^2
\ene
to transform, respectively, the focusing mKdV equation~\cite{miura,soliton}
\bee\label{mKdV}
 u_t+6u^2u_x+u_{xxx}=0
\ene
and the defocusing mKdV equation~\cite{miura,soliton}
\bee\label{mKdV2}
 u_t-6u^2u_x+u_{xxx}=0
\ene
into the same KdV equation~\cite{kdv}
\bee\label{KdV}
 v_t+6vv_x+v_{xxx}=0,
\ene
which can describe the shallow water wave, pressure waves,  acoustic waves,  magneto-sonic waves,  electron plasma waves,
and ion acoustic waves~\cite{Miura76,Schamel73}.

In this subsection, the data-driven deep learning method will be used in two different cases. In the first part, the parameters of Miura transform will be discovered including with and without disturbance terms. The data-set is generated by an exact soliton solution of the mKdV equation and an corresponding exact solution of the KdV equation, which is generated by following soliton solution through the Miura transform. In the second part, an equation will be discovered through the solution of another equation. The data-set is generated by the solution of another equation. The real Miura transform will be used in the T-part loss function to find the unknown equation. In the both cases, the F-part loss is obtained by the exact solution of the mKdV equation given by Eq.~(\ref{mKdV-solu}) or (\ref{mKdV-solu1}) and the original mKdV equation given by Eq.~(\ref{mKdV}) or (\ref{mKdV2}). And the G-part loss is constructed from the exact solution of the KdV equation given by Eq.~(\ref{KdV-solu}) or (\ref{KdV-solu1}) and the original KdV equation (\ref{KdV}).

\begin{table}[!t]
	\centering
    \setlength{\tabcolsep}{16pt}% column separation
    \renewcommand{\arraystretch}{1.4}%row space
	\caption{Data-driven discovery of the complex Miura transform (\ref{miura4}) via system (\ref{miura-sys}): $a$, $b$, $c,\, d$ and their errors, as well as the training times. \vspace{0.05in}}
	\begin{tabular}{cccccc} \hline\hline
	Case & $a$  & $b$  & $c$  & $d$  &  \\  \hline
  Exact & 1 & 1 &  0 &0 &  \\
  A (no noise) & 0.99931 & 0.99958& 0 & 0 & \\
  A ($2\%$ noise) & 1.00026 & 0.99981 & 0 & 0 &  \\
   B (no noise) &  1.00006 &  0.99990 & -4.1$\times10^{-4}$ & -1.0$\times10^{-5}$ &  \\
   B ($2\%$ noise) & 1.00017 & 0.99994 & -1.3$\times10^{-4}$ & 2.0$\times10^{-4}$ &  \\
    \hline\hline
   	Case & error of $a$ & error of $b$ & error of $c$ & error of $d$ & time \\  \hline
    A (no noise) & 6.94$\times10^{-4}$ & 4.25$\times10^{-4}$ & 0 & 0 &  609.30s \\
  A ($2\%$ noise) & 2.58$\times10^{-4}$ & 1.89$\times10^{-4}$ & 0 & 0 &  645.68s \\
   B (no noise)  & 6.00$\times10^{-5}$ & 1.03$\times10^{-4}$ & 4.05$\times10^{-4}$  & 5.00$\times10^{-6}$ &  635.56s \\
   B ($2\%$ noise) & 1.68$\times10^{-4}$  & 5.90$\times10^{-5}$  & 1.32$\times10^{-4}$ & 1.99$\times10^{-5}$ & 637.18s \\
    \hline\hline
	\end{tabular}
	\label{miura-table1}
\end{table}

\v {\bf  Case 1.\, Data-driven discovery of the complex Miura transform} \v

In the following, we would like to study the data-driven parameter discovery of the complex Miura transform. We consider the generalized Miura transform
\bee\label{miura4}
 v=iau_x+bu^2+cuu_x+du_{xx},
\ene
where $a,b,c,d$ are four parameters to be determined latter. If $a=b=1,\, c=d=0$, then the transform (\ref{miura4}) reduces to the known Miura transform (\ref{miura}).

In what follows, we would like to use the above-mentioned deep leaning scheme to discover these parameters $a,\,b,\,c,\,d$  of the complex Miura transform (\ref{miura4}) between the focusing mKdV equation and KdV equation in two cases by considering the system
\bee\label{miura-sys}
\left\{\begin{array}{l}
 u_t+6u^2u_x+u_{xxx}=0, \vspace{0.05in} \\
  v_t+6vv_x+v_{xxx}=0, \vspace{0.05in}\\
  v-iau_x-bu^2-cuu_x-du_{xx}=0.
   %\vspace{0.05in}\\
  %u(x,0)=u_0,\quad u(-L, x)=u(L,x),\vspace{0.05in}\\
  %v(x,0)=v_0(x),\quad v(-L, x)=v(L,x)
\end{array}\right.
\ene
 The training data-set is generated from the known bright soliton of the focusing mKdV equation (\ref{mKdV}):
\bee\label{mKdV-solu}
 u=k {\rm sech}(k x-k^3t+x_0),
\ene
where $k$ is a non-zero free real parameter, and $x-0$ is arbitrary real constant. And the corresponding complex bright soliton of the KdV equation is
\bee\label{KdV-solu}
 v=k^2 {\rm sech}^2(k x-k^3t+x_0)-ik^2 {\rm sech}(k x-k^3t+x_0)\tanh(kx-k^3t+x_0)
\ene
by the complex Miura transform (\ref{miura}).

\begin{figure}[!t]
\hspace{2.5in}
\begin{center}
 \hspace{0.2in}{\scalebox{0.75}[0.65]{\includegraphics{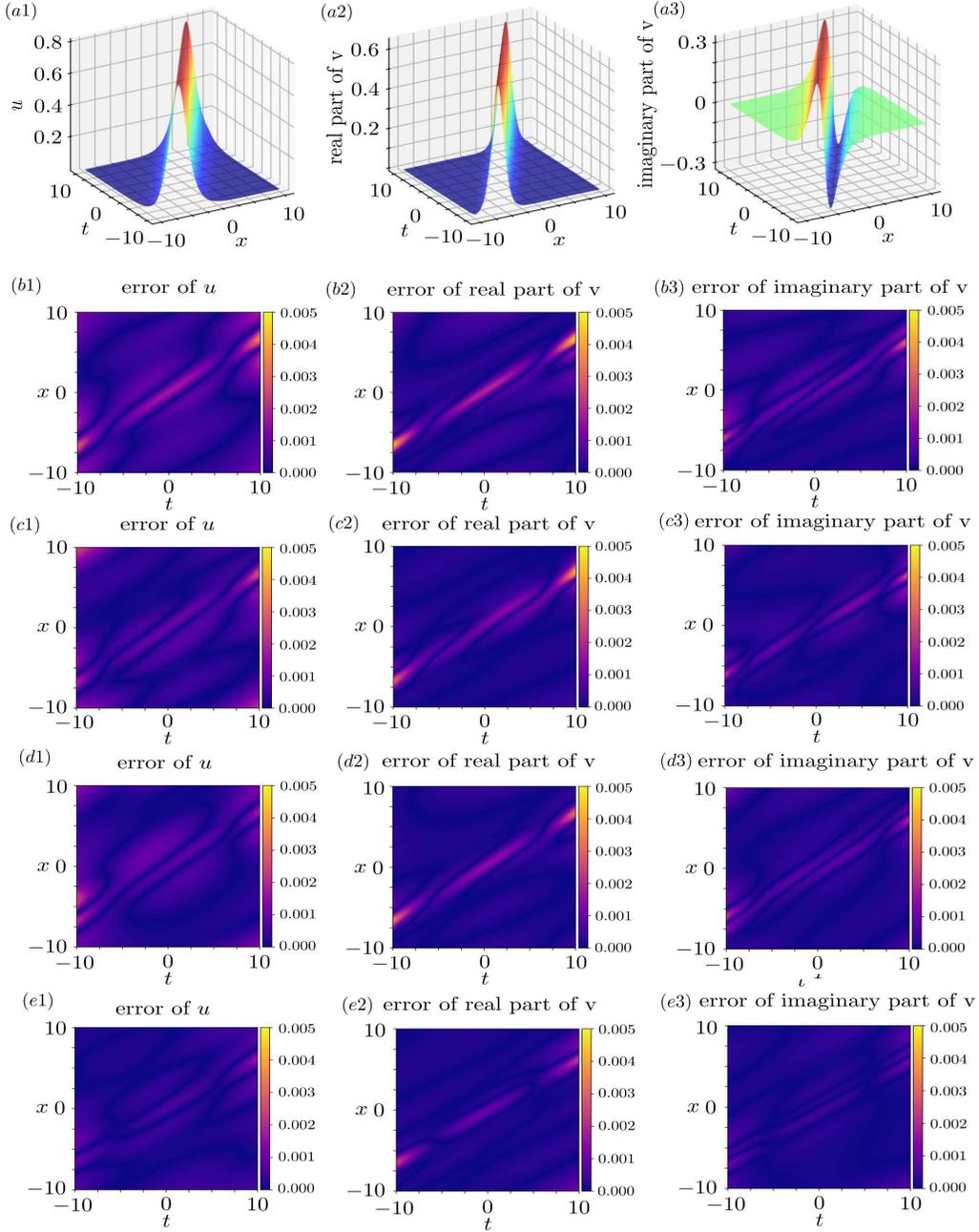}}}
\end{center}
\par
\vspace{-0.05in}
\caption{\small Data-driven complex Miura transform discovery. (a1) soliton (\ref{mKdV-solu}), (a2, a3) real and imaginary parts of soliton (\ref{KdV-solu}). (b1-e3) Relative errors between exact solutions and neural network solutions: (b1-b3) Case A without a noise, (c1-c3) Case A with a 2$\%$ noise, (d1-d3) Case B without a noise,
(e1-e3) Case B with a 2$\%$ noise. The relative $\mathbb{L}^2-$norm errors of $u(x,t)$, ${\rm Re}(v(x,t))$ and ${\rm Im}(v(x,t))$ are (b1) 2.11e-3, (b2) 2.71e-3, (b3) 2.68e-3, (c1) 1.69e-3, (c2) 2.17e-3, (c3) 2.16e-3, (d1) 1.58e-3, (d2)
2.00e-3, (d3) 2.17e-3, (e1) 9.43e-3, (e2) 1.53e-3, and (e3) 1.43e-3, respectively.}
\label{fig3-learnmiura-foc}
\end{figure}

\v{\it Case A}.---We fix $c=d=0$, and learn the two unknown parameters $a,\, b$. The data-set is sampled in the spatio-temporal region $(x,t)\in [-10,10]\times[-10,10]$. Moreover, 10,000 sampling points will be used in the training process, and $k=0.8$ in this example. A 6-layer neural network with 40 neurons per layer is used to learn system (\ref{miura-sys}) to fit the exact solutions of two equations. For convenience, the initial values of all free parameters are set as 1. We choose 10,000 steps Adam and 20,000 steps L-BFGS optimizations to train the considered deep learning model.
Case A in Table~\ref{miura-table1} exhibits the learning parameters about $a,\, b$, and their errors under two senses of the training data without a noise and with a $2\%$ noise, which imply that the used deep learning method is effective.
%Moreover, the errors are exhibited in Figs.~\ref{fig2-SGerror}(b1, b2) for the cases without a noise and with a $2\%$ noise, respectively.
The training times are 609.30s and 645.68s,, respectively.  Moreover, the errors are exhibited in Figs.~\ref{fig3-learnmiura-foc}(b1-b4) for the cases without a noise and with a $2\%$ noise, respectively.

\v{\it Case B}.---We learn all four unknown parameters $a,\, b,\, c,\, d$.   We used the same deep neural network method as Case A to study this case. Case B in Table~\ref{miura-table1} displays the learning parameters about $a,\,b,\,c,\, d$, and their errors under two senses of the training data without a noise and with a $2\%$ noise, which imply that the used deep learning method is effective.
 %Moreover, the errors are exhibited in Figs.~\ref{fig2-SGerror}(b3, b4) for the cases without a noise and with a $2\%$ noise, respectively.
 The training times are 635.56s and 637.18s, respectively. Moreover, the errors are exhibited in Figs.~\ref{fig3-learnmiura-foc}(c1-c4) for the cases without a noise and with a $2\%$ noise, respectively.

\begin{table}[!t]
	\centering
    \setlength{\tabcolsep}{15pt}% column separation
    \renewcommand{\arraystretch}{1.4}%row space
	\caption{Data-driven discovery of real Miura transform (\ref{miura5}) via system (\ref{miura-sys2}): $a$, $b$, $c,\, d$ and their errors, as well as the training times. \vspace{0.05in}}
	\begin{tabular}{cccccc} \hline\hline
	Case & $a$ &  $b$ &  $c$ &  $d$ &  \\  \hline
  Exact & 1  & -1 & 0 & 0 &  \\
  Case A (no noise) & 0.99999 &  -1.00000 & 0 & 0 &  \\
  Case A ($2\%$ noise) & 1.00026 & -1.00012 & 0 & 0 &  \\
  Case B (no noise) & 1.00000 & -0.99999 &  0.00011 & -0.00013 &  \\
  Case B ($2\%$ noise) & 1.00019 & -1.00023 &  0.00271 & -0.00303 &  \\
   \hline\hline
   Case & error of $a$ & error of $b$ & error of $c$ & error of $d$ & time \\  \hline
   Case A (no noise) &  1.36$\times10^{-5}$ & 1.35$\times10^{-6}$ & 0 & 0 &  374.40s\\
  Case A ($2\%$ noise) & 2.55$\times10^{-4}$ & 1.22$\times10^{-4}$ & 0 & 0 &  407.44s\\
  Case B (no noise) & 7.39$\times10^{-6}$  & 5.66$\times10^{-6}$  & 1.14$\times10^{-4}$ & 1.25$\times10^{-4}$ &405.32s\\
  Case B ($2\%$ noise) & 1.85$\times10^{-4}$  & 2.31$\times10^{-5}$ & 2.71$\times10^{-3}$ &  3.03$\times10^{-3}$ &417.00s \\
     \hline\hline
	\end{tabular}
	\label{miura-table2r}
\end{table}

\v{\bf Case 2.\, Data-driven discovery of the real Miura transform} \v

We consider the generalized form of the real Miura transform (\ref{miura2}) as
\bee\label{miura5}
 v=au_x+bu^2+cuu_x+du_{xx}.
\ene
where $a,\,b,\,c,\,d$ are four real parameters to be determined. If $a=b=1,\, c=d=0$, then the transform (\ref{miura5}) reduces to the known real Miura transform (\ref{miura2}).

In what follows, we would like to use the above-mentioned deep leaning scheme to discover these parameters $a,\,b,\,c,\,d$ of the Miura transform (\ref{miura5}) between the mKdV equation and KdV equation in two cases by considering the system
\bee\label{miura-sys2}
\left\{\begin{array}{l}
 u_t-6u^2u_x+u_{xxx}=0, \vspace{0.05in} \\
  v_t+6vv_x+v_{xxx}=0, \vspace{0.05in}\\
  v-au_x-bu^2-cuu_x-du_{xx}=0.
   %\vspace{0.05in}\\
  %u(x,0)=u_0,\quad u(-L, x)=u(L,x),\vspace{0.05in}\\
  %v(x,0)=v_0(x),\quad v(-L, x)=v(L,x)
\end{array}\right.
\ene
 The training data-set is obtained through a shock wave solution of the defocusing mKdV equation (\ref{mKdV2})
\bee\label{mKdV-solu1}
 u(x,t)=k\tanh(k x+k^3t)
\ene
with a free real parameter $k\not=0$, and the soliton solution of the KdV equation (\ref{KdV})
\bee\label{KdV-solu1}
 v(x,t)=2k^2{\rm sech}^2(kx+2k^3t)-k^2
\ene
via the real Miura transform (\ref{miura2}).

\begin{figure}[!t]
\hspace{2.5in}
\begin{center}
\hspace{0.3in} {\scalebox{0.75}[0.75]{\includegraphics{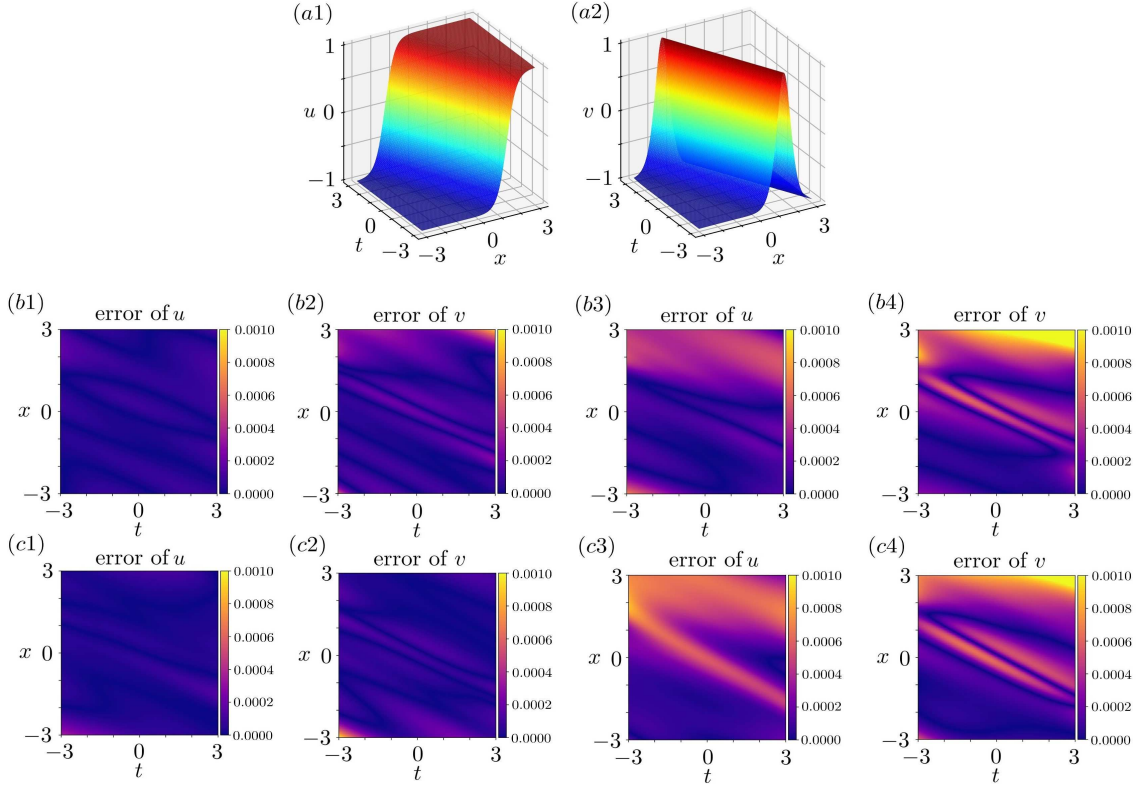}}}
\end{center}
\par
\vspace{-0.05in}
\caption{\small Data-driven discovery of real Miura transform. (a1) solution(\ref{mKdV-solu1}),(a2) soliton solution(\ref{KdV-solu1}).
(b1-e4) Relative errors between exact solutions and neural network solutions:
(b1-b2) Case A without a noise, (b3-b4) Case A with a 2$\%$ noise,(c1-c2) Case B without a noise, (c3-c4)
Case B with a 2$\%$ noise. The relative $\mathbb{L}^2-$norm errors of $u(x,t)$ and $v(x,t)$ are (b1) 5.74e-5,
(b2) 1.44e-4, (b3) 2.69e-4, (b4) 4.42e-4, (c1) 5.37e-5, (c2) 9.37e-5, (c3) 4.17e-4, and (c4) 3.91e-4, respectively.}
\label{fig4-learnmiura-defoc}
\end{figure}

\v{\it Case A}.---We fix $c=d=0$, and learn the two parameters $a,\, b$. The data-set is sampled in the spatio-temporal region $(x,t)\in [-3,3]\times[-3,3]$. Moreover, 10,000 sampling points will be used in the training process, and $k=1$ in this example. A 7-layer neural network with 20 neurons per layer is used to learn system (\ref{miura-sys2}) to fit the exact solutions of two equations. For convenience, the initial value of all free parameters are set as 1. We choose 5,000 steps Adam and 5,000 steps L-BFGS optimizations to train the considered deep learning model.
Case A in Table~\ref{miura-table2r} exhibits the learning parameters about $a,\, b$, and their errors under two senses of the training data without a noise and with a $2\%$ noise, which imply that the used deep learning method is effective.
%Moreover, the errors are exhibited in Figs.~\ref{fig2-SGerror}(b1, b2) for the cases without a noise and with a $2\%$ noise, respectively.
The training times are  374.40s and 407.44s, respectively.  Moreover, the errors are exhibited in Figs.~\ref{fig4-learnmiura-defoc}(b1-c3) for the cases without a noise and with a $2\%$ noise, respectively.

\v{\it Case B}.---We learn all four parameters $a,\, b,\, c,\, d$.   We used the same deep neural network method as Case A to study this case. Case B in Table~\ref{miura-table2r} displays the learning parameters about $a,\,b,\,c,\, d$, and their errors under two senses of the training data without a noise and with a $2\%$ noise, which imply that the used deep learning method is effective.
 %Moreover, the errors are exhibited in Figs.~\ref{fig2-SGerror}(b3, b4) for the cases without a noise and with a $2\%$ noise, respectively.
 The training times are 405.32s and 417.00s, respectively. Moreover, the errors are exhibited in Figs.~\ref{fig4-learnmiura-defoc}(d1-e3) for the cases without a noise and with a $2\%$ noise, respectively.

\subsection{Data-driven discoveries of mKdV equations via Miura transforms}\label{viamu}

In the following we will use the known complex Miura transform (\ref{miura}) or real Miura transform (\ref{miura2}) and the corresponding known solution (\ref{KdV-solu}) or (\ref{KdV-solu1}) of the KdV equation (\ref{KdV}) to discover the unknown mKdV equation given by Eq.~(\ref{mKdV3g}) by using the above deep learning scheme.

\subsubsection{Data-driven discovery of the focusing mKdV equation via the complex Miura transform}

In this subsection, we would like to use the above-mentioned deep learning scheme to learn the mKdV equation through the complex Miura transform (\ref{miura}). The solution $u$ of the focusing mKdV equation (\ref{mKdV}) can not be explicitly expressed by the solution $v$ of the KdV equation (\ref{KdV}). The training data-set of $u$ can not be directly calculated by the data-set of $v$. So, we will use a neural network to approximate $u$ directly. Training the loss ${\rm TL}_T$ brings $u$ close to the right solution.

We would like to consider the generalized form of the original mKdV equation (\ref{mKdV})
\bee\label{mKdV3g}
 u_t+au^2u_x+bu_{xxx}+cuu_{xx}+du^4=0
\ene
to test the robustness of our scheme, where $a$, $b$, $c$, $d$ are real-valued parameters to be determined.

We use the above-mentioned deep leaning scheme to discover the parameters of the mKdV equation (\ref{mKdV3g}) via the complex Miura transform (\ref{miura}) and KdV equation (\ref{KdV}) in two cases by considering the system
\bee\label{miura-kdv-sys}
\left\{\begin{array}{l}
  u_t+au^2u_x+bu_{xxx}+cuu_{xx}+du^4=0, \vspace{0.05in} \\
  %v_t+6vv_x+v_{xxx}=0, \vspace{0.05in}\\
  v-iu_x-u^2=0.
   %\vspace{0.05in}\\
  %u(x,0)=u_0,\quad u(-L, x)=u(L,x),\vspace{0.05in}\\
  %v(x,0)=v_0(x),\quad v(-L, x)=v(L,x)
\end{array}\right.
\ene

\begin{figure}[!t]
%\hspace{2.5in}
\begin{center}
 \hspace{0.2in} {\scalebox{0.75}[0.75]{\includegraphics{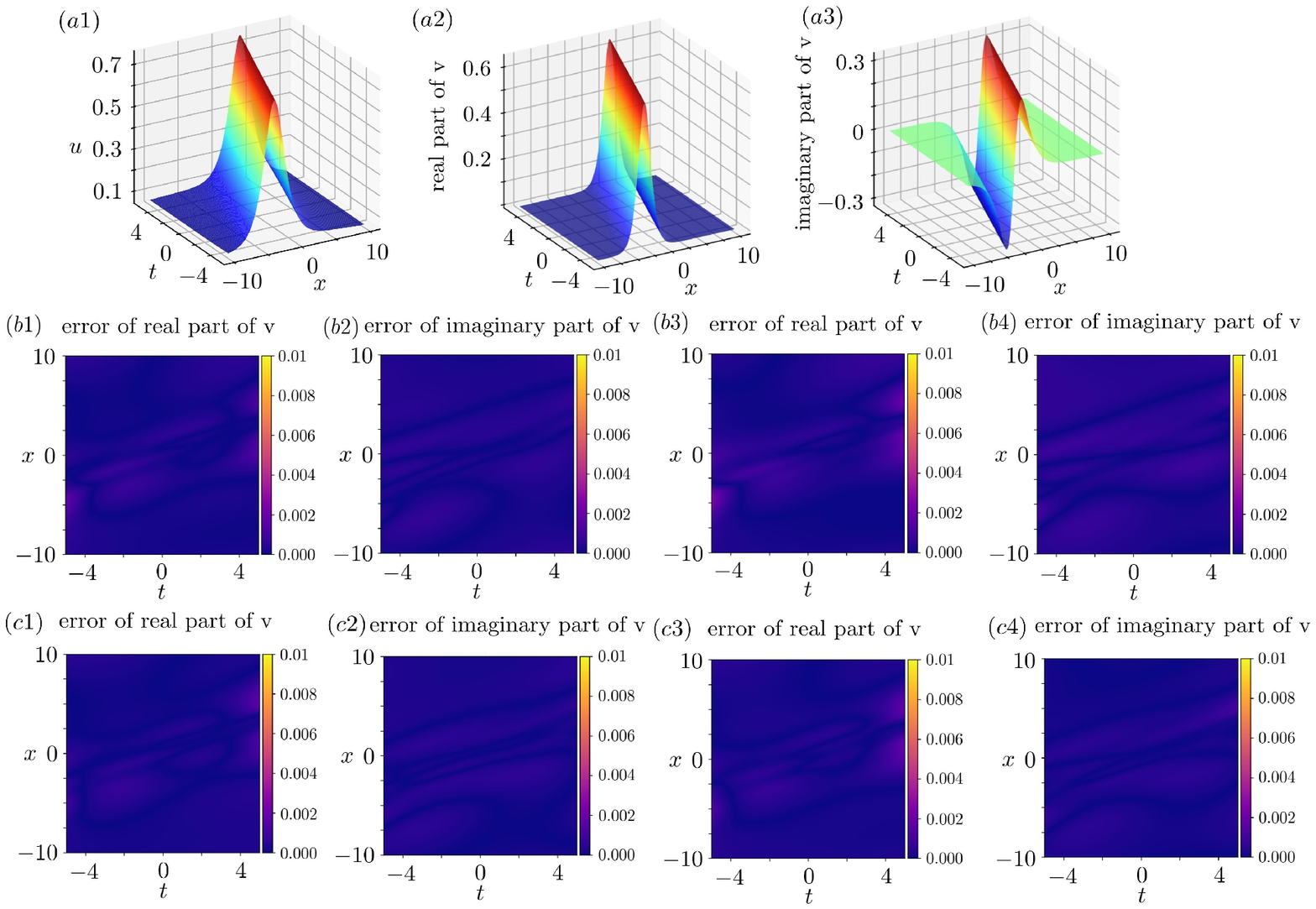}}}
\end{center}
\par
\vspace{-0.05in}
\caption{\small Data-driven discovery of the focusing mKdV equation via the complex Miura transform. (a1) trained soliton of mKdV equation,
(a2-a3) real and imaginary parts of trained soliton of the KdV equation. (b1, b2) Case A without a nose,
(b3, b4) Case A with a 2$\%$ noise, (c1, c2) Case B without a noise, (c3, c4) Case B with a 2$\%$ noise.
The relative $\mathbb{L}^2-$norm errors of ${\rm Re}(v(x,t))$ and ${\rm Im}(v(x,t))$ are (b1) $1.66\times 10^{-3}$, (b2) $2.30\times 10^{-3}$, (b3)$2.26\times 10^{-3}$, (b4)$3.27\times 10^{-3}$, (c1)$1.49\times 10^{-3}$, (c2)$2.03\times 10^{-3}$, (c3) $1.68\times 10^{-3}$, and (c4)$1.92\times 10^{-3}$, respectively.}
\label{fig5-learnmKdV-foc}
\end{figure}

\v{\it Case A}.---We fix $c=d=0$, and learn the two parameters $a,\, b$. The training data-set is generated by the soliton (\ref{KdV-solu}) of the KdV equation. The data-set is sampled in the spatio-temporal region $(x,t)\in [-10,10]\times[-5,5]$. Moreover, 10,000 sampling points will be used in the training process, and $k=0.8$ in this example. A 6-layer neural network with 40 neurons per layer is used to learn system (\ref{miura-kdv-sys}) to fit the exact solutions of two equations. For convenience, the initial values of all free parameters are chosen as 1. We choose 10,000 steps Adam and 20,000 steps L-BFGS optimizations to train the considered deep learning model.
Case A in Table~\ref{miura-table3} exhibits the learning parameters about $a,\, b$, and their errors under two senses of the training data without a noise and with a $2\%$ noise, which imply that the used deep learning method is effective.
%Moreover, the errors are exhibited in Figs.~\ref{fig2-SGerror}(b1, b2) for the cases without a noise and with a $2\%$ noise, respectively.
The training times are  768.54s and 764.55s, respectively. Moreover, the errors are exhibited in Figs.~\ref{fig5-learnmKdV-foc}(b1-b4) for the cases without a noise and with a $2\%$ noise, respectively.

\v{\it Case B}.---We learn all four parameters $a,\, b,\, c,\, d$. We used the same deep neural network method as Case A to study this case. Case B in Table~\ref{miura-table3} displays the learning parameters about $a,\,b,\,c,\, d$, and their errors under two senses of the training data without a noise and with a $2\%$ noise, which imply that the used deep learning method is effective.
 %Moreover, the errors are exhibited in Figs.~\ref{fig2-SGerror}(b3, b4) for the cases without a noise and with a $2\%$ noise, respectively.
 The training times are 742.11s and 750.01s, respectively. Moreover, the errors are exhibited in Figs.~\ref{fig5-learnmKdV-foc}(c1-c4) for the cases without a noise and with a $2\%$ noise, respectively.

\begin{table}[!t]
	\centering
    \setlength{\tabcolsep}{16pt}% column separation
    \renewcommand{\arraystretch}{1.4}%row space
	\caption{Data-driven discovery of the focusing mKdV equation (\ref{mKdV3g}) via system (\ref{miura-kdv-sys}) with the complex Miura transform (\ref{miura}): $a, b, c, d$, and their errors, as well as the training times. \vspace{0.05in}}
	\begin{tabular}{cccccc} \hline\hline
	Case & $a$  & $b$ & $c$  & $d$  & \\  \hline
    Exact & 6 & 1 & 0 & 0 &  \\
  A (no noise) & 5.94175 & 0.98650  & 0 & 0 &  \\
  A ($2\%$ noise) & 5.90499 & 0.97883 & 0 & 0 &  \\
   B (no noise) & 5.96718 & 0.99260  & 1.88$\times10^{-3}$ & 1.37$\times10^{-3}$ &  \\
   B ($2\%$ noise) & 5.95149  & 0.98922 & 3.3$\times10^{-4}$ &  -3.8$\times10^{-4}$ & \\
    \hline\hline
  Case & error of $a$ & error of $b$ & error of $c$ & error of $d$ & time \\  \hline
    A (no noise) & 5.83$\times10^{-2}$ & 1.35$\times10^{-2}$ & 0 & 0 & 768.54s \\
  A ($2\%$ noise) & 9.50$\times10^{-2}$ & 2.12$\times10^{-2}$ & 0 & 0 & 764.55s \\
   B (no noise) & 3.28$\times10^{-2}$ & 7.40$\times10^{-3}$ & 1.88$\times10^{-3}$ & 1.37$\times10^{-3}$ & 742.11s \\
   B ($2\%$ noise) & 4.85$\times10^{-2}$ & 1.08$\times10^{-2}$ & 3.27$\times10^{-4}$ & 3.80$\times10^{-4}$ &750.01s \\
    \hline\hline
	\end{tabular}
	\label{miura-table3}
\end{table}

\subsubsection{Data-driven discovery of the  defocusing mKdV equation via the real Miura transform}

In this subsection, we would like to train the defocusing mKdV equation through the real Miura transform (\ref{miura2}). In general, the solution $u$ of the  defocusing mKdV equation (\ref{mKdV2}) is very difficultly expressed by the solution $v$ of the KdV equation (\ref{KdV}). The training data of $u$ can not be directly generated by the data-set of $v$. So, we will use a neural network to approximate $u$ directly. The smaller training loss ${\rm TL}_T$ can make the trained $\hat{u}$ close to real solution $u$.

We would like to consider the generalized form of the original mKdV equation given by Eq.~(\ref{mKdV3g}) to test the robustness of our scheme. We use the above-mentioned deep leaning scheme to discover the parameters of the mKdV equation (\ref{mKdV3g}) via the real  Miura transform (\ref{miura2}) and KdV equation (\ref{KdV}) in two cases by considering the system
\bee\label{miura-kdv-sys2}
\left\{\begin{array}{l}
  u_t+au^2u_x+bu_{xxx}+cuu_{xx}+du^4=0, \vspace{0.05in} \\
  %v_t+6vv_x+v_{xxx}=0, \vspace{0.05in}\\
  v-u_x+u^2=0.
   %\vspace{0.05in}\\
  %u(x,0)=u_0,\quad u(-L, x)=u(L,x),\vspace{0.05in}\\
  %v(x,0)=v_0(x),\quad v(-L, x)=v(L,x)
\end{array}\right.
\ene

\begin{figure}[!t]
\hspace{2.5in}
\begin{center}
 \hspace{0.2in} {\scalebox{0.75}[0.75]{\includegraphics{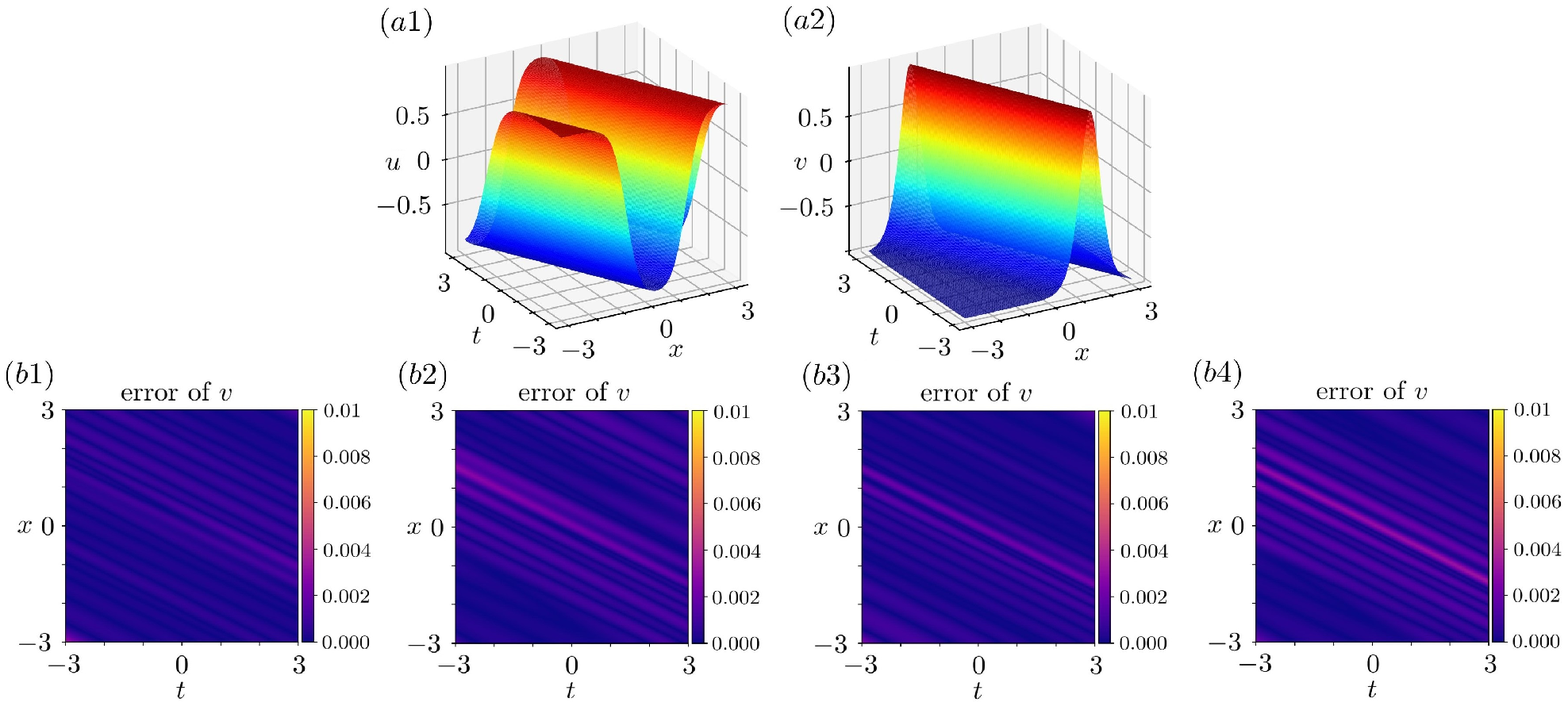}}}
\end{center}
\par
\vspace{-0.05in}
\caption{\small Data-driven discovery of the defocusing mKdV equation via the real Miura transform. (a1) trained soliton of mKdV equation,
(a2) trained soliton of the KdV equation. (b1) Case A without a nose, (b2) Case A with a 2$\%$ noise, (b3) Case B without a noise, (b4) Case B with a 2$\%$ noise. The relative $\mathbb{L}^2-$norm errors of $v(x,t)$ are (b1)$5.44\times 10^{-4}$, (b2)$8.25\times 10^{-4}$, (b3)$6.19\times 10^{-4}$, and (b4)$9.64\times 10^{-4}$, respectively.  }
\label{fig6-learnmKdV-defoc}
\end{figure}

\v{\it Case A}.---We fix $c=d=0$, and learn the two unknown parameters $a,\, b$. The training data-set is generated by the soliton (\ref{KdV-solu1}) of the KdV equation. The data-set is sampled in the spatio-temporal region $(x,t)\in [-3,3]\times[-3,3]$. Moreover, 10,000 sampling points will be used in the training process, and $k=0.8$ in this example. A 7-layer neural network with 20 neurons per layer is used to learn system (\ref{miura-kdv-sys2}) to fit the exact solutions of two equations. For convenience, the initial value of all free parameters are set as 1. We choose 5,000 steps Adam and 5,000 steps L-BFGS optimizations to train the considered deep learning model.
Case A in Table~\ref{miura-table4} exhibits the learning parameters about $a,\, b$, and their errors under two senses of the training data without a noise and with a $2\%$ noise, which imply that the used deep learning method is effective.
%Moreover, the errors are exhibited in Figs.~\ref{fig2-SGerror}(b1, b2) for the cases without a noise and with a $2\%$ noise, respectively.
The training times are  203.41s  and 178.44s, respectively. Moreover, the errors are exhibited in Figs.~\ref{fig6-learnmKdV-defoc}(b1, b2) for the cases without a noise and with a $2\%$ noise, respectively.

\v{\it Case B}.---We learn all four unknown parameters $a,\, b,\, c,\, d$. We used the same deep neural network method as Case A to study this case. Case B in Table~\ref{miura-table4} displays the learning parameters about $a,\,b,\,c,\, d$, and their errors under two senses of the training data without a noise and with a $2\%$ noise, which imply that the used deep learning method is effective.
 %Moreover, the errors are exhibited in Figs.~\ref{fig2-SGerror}(b3, b4) for the cases without a noise and with a $2\%$ noise, respectively.
 The training times are 178.08s and 163.98s, respectively. Moreover, the errors are exhibited in Figs.~\ref{fig6-learnmKdV-defoc}(b3, b4) for the cases without a noise and with a $2\%$ noise, respectively.

\begin{table}[!t]
	\centering
    \setlength{\tabcolsep}{12pt}% column separation
    \renewcommand{\arraystretch}{1.4}%row space
	\caption{Data-driven  defocusing mKdV equation (\ref{mKdV3g}) discovery via system (\ref{miura-kdv-sys2}) with the real Miura transform (\ref{miura2}): $a, b, c, d$, and their errors, as well as the training times. \vspace{0.05in}}
	\begin{tabular}{cccccc} \hline\hline
	Case & $a$ & $b$ & $c$ & $d$  &  \\  \hline
  Exact & -6 & 1 & 0 & 0 &  \\
  Case A (no noise) & -5.98709 & 0.99742 & 0 & 0 &  \\
  Case A ($2\%$ noise) & -5.97933 &  0.99527 & 0 & 0 &  \\
  Case B (no noise) & -5.98739 & 0.99681 & -0.00330 & -0.00122 &  \\
  Case B ($2\%$ noise) & -5.97631 & 0.99433 & -0.00204 & -0.00149 &  \\
   \hline\hline
  Case & error of $a$ & error of $b$ & error of $c$ & error of $d$ & time \\  \hline
   Case A (no noise) & 1.29$\times10^{-2}$ & 2.58$\times10^{-3}$ & 0 & 0 & 203.41s \\
  Case A ($2\%$ noise) &  2.07$\times10^{-2}$ & 4.73$\times10^{-3}$ & 0 & 0 &  178.44s \\
  Case B (no noise)  & 1.26$\times10^{-2}$  & 3.19$\times10^{-3}$ & 3.30$\times10^{-3}$ & 1.22$\times10^{-3}$ & 178.08s \\
  Case B ($2\%$ noise) & 2.37$\times10^{-2}$ & 5.67$\times10^{-3}$ & 2.04$\times10^{-3}$ & 1.50$\times10^{-3}$ & 163.98s \\
      \hline\hline
	
\end{tabular}
	\label{miura-table4}
\end{table}

\section{The BT-enhanced scheme for the data-driven PDE discovery}

Some deep learning schemes about the PDE discovery usually only used the basic physical data (e.g., PDE solution data) and the possible forms of the presupposed PDEs (see, e.g., Ref.~\cite{raiss19}). Sometime they are not effective. In this section, we will propose an enhanced scheme of PDE discovery based on the BTs. We can consider the BTs into two forms, including implicit and explicit forms, to learn the soliton equations.

\subsection{The BT-enhanced deep learning scheme for the PDE discovery}

In this section, we will provide the framework of our scheme. The main idea about the PDE discovery in the previous works is to
use the solution information, which may be generated by the experimental data-set or given one.
But if we know the BT of the original equation, then we can find the higher accuracies of parameters in the original equation.
The basic idea of this method is to add the high-order constraints generated by the BTs to the training process.

We will discuss our scheme in the two different cases. In the first case, the BT can be written as an explicit form. In the second case, the BT can be given only in an implicit form.

\begin{figure}[!t]
\hspace{2.5in}
\begin{center}
 \hspace{0.3in} {\scalebox{0.65}[0.65]{\includegraphics{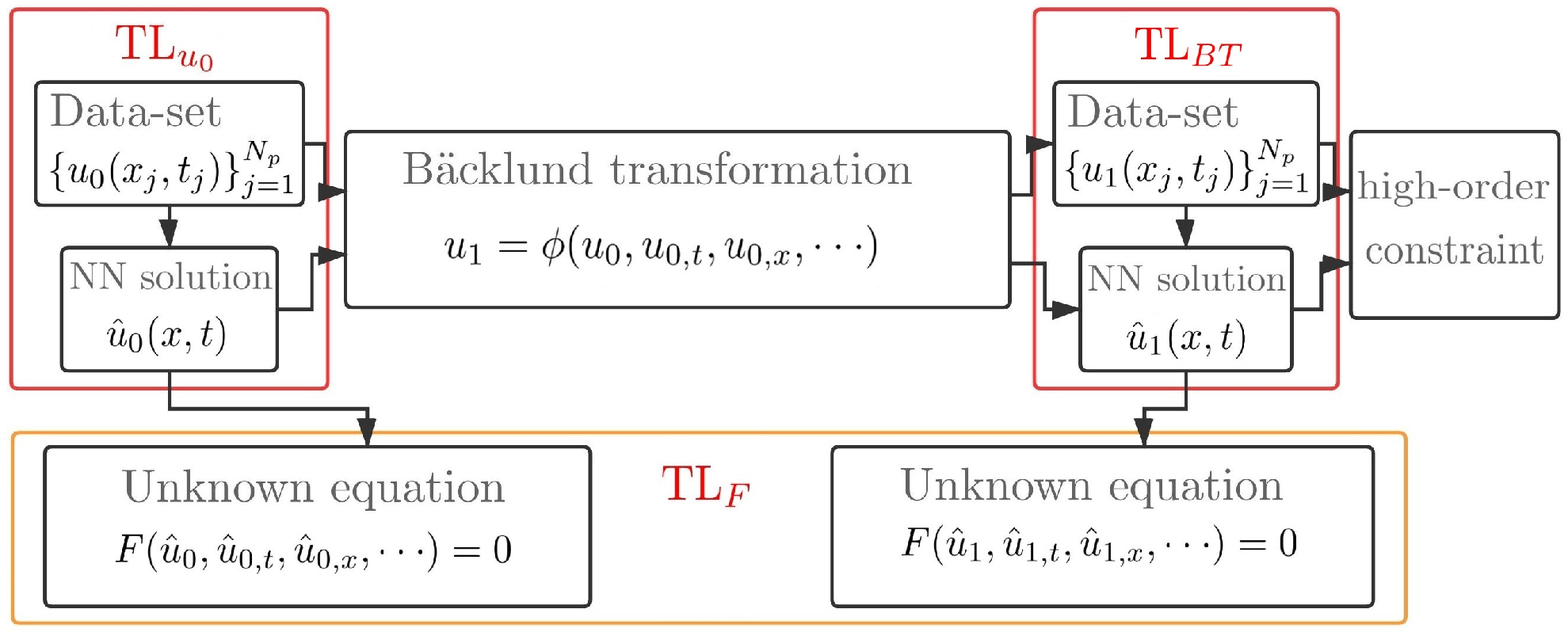}}}
\end{center}
\par
\vspace{0.05in}
\caption{\small The BT-enhanced PDE discovery scheme for the explicit BTs.}
\label{fig7-DNN2}
\end{figure}

\v {\it Case 1.}---Fig~\ref{fig7-DNN2} exhibits the case that the BT can be written as an explicit form.
$\hat{u}_0(x,t)$ and $\hat{u}_1(x,t)$ are represented by some neural networks, such as the fully-connected neural networks
or other kinds of neural networks. The data-set $\{u_0(x_j,t_j)\}_{j=1}^{N_p}$ arises from the experimental data or solution data. And the first-order data-set $\{u_1(x_j,t_j)\}_{j=1}^{N_p}$ is only calculated from the above data-set $\{u_0(x_j,t_j)\}_{j=1}^{N_p}$ by some BT. The unknown equations contain some parameters to be learned. The train loss contains
the three parts: ${\rm TL}_{u_0}$, ${\rm TL}_{BT}$ and ${\rm TL}_{F}$, where  ${\rm TL}_{u_0}$ and ${\rm TL}_{u_1}$ make the neural network model fit the data-set, and
the ${\rm TL}_{F}$ makes the unknown equation approach the exact form, which composed by the above neural network solution. If necessary, the high-order solutions can generate the stronger constraints. It is worth noting that the data-set $\{u_1(x_j,t_j)\}_{j=1}^{N_p}$ should be prepared before the training.

%We rewrite $u_k=v_k+iw_k$ and $\hat{u}_k=\hat{v}_k+i\hat{w}_k,\, (k=0,1)$, and the unknown equation $F$ as
%$F=F_1+iF_2$.
The train loss is given by  $ {\rm TL}={\rm TL}_{u_0}+{\rm TL}_{BT}+{\rm TL}_{F},$
where
\bee
 &&{\rm TL}_{u_0}= \displaystyle\frac{1}{N_p}\sum_{j=1}^{N_p}\left|u_0(x_j,t_j)-\hat{u}_0(x_j,t_j)\right|^2,\v\\
  &&{\rm TL}_{BT}= \displaystyle\frac{1}{N_p}\sum_{j=1}^{N_p}\left|BT(u_0(x_j,t_j))-\hat{u}_1(x_j,t_j)\right|^2,\v\\
 &&{\rm TL}_{F}= \frac{1}{N_p}\sum_{j=1}^{N_p}\left(
 \left|F(\hat{u}_0(x_j,t_j))\right|^2+ \left|F(\hat{u}_1(x_j,t_j))\right|^2\right),
\ene
in which ${\rm TL}_{u_0}$ and ${\rm TL}_{BT}$ make the neural network model fit the data-set, ${\rm TL}_{F}$ makes the unknown equation approach to the exact form, which composed by above neural network solutions $\hat{u}_0$ and $\hat{u}_1$. If necessary, high-order solution can be used to obtain the stronger constraints. For instance, if we use $\{u_0(x_j,t_j)\}_{j=1}^{N_p}$, $\{BT(u_0(x_j,t_j))\}_{j=1}^{N_p}$ and 2-order data-set $\{BT^2(u_0(x_j,t_j))\}_{j=1}^{N_p}$ to train our model to obtain
the neural network solutions $\hat{u}_j (j=0, 1, 2)$. The losses ${\rm TL}_{BT}$ and  ${\rm TL}_{F}$ in the TL are replaced  by
\bee
{\rm TL}_{BT^2}=\displaystyle\frac{1}{N_p}\sum_{j=1}^{N_p}\left(
\left|BT(u_0(x_j,t_j))-\hat{u}_1(x_j,t_j)\right|^2+\left|BT^2(u_0(x_j,t_j))-\hat{u}_2(x_j,t_j)\right|^2 \right),
\ene
\bee
{\rm TL}_{F}=\d \frac{1}{N_p}\sum_{j=1}^{N_p}\left(
 \left|F(\hat{u}_0(x_j,t_j))\right|^2+ \left|F(\hat{u}_1(x_j,t_j))\right|^2+ \left|F(\hat{u}_2(x_j,t_j))\right|^2\right).
\ene

\begin{figure}[!t]
\begin{center}
 \hspace{0.35in} {\scalebox{0.75}[0.75]{\includegraphics{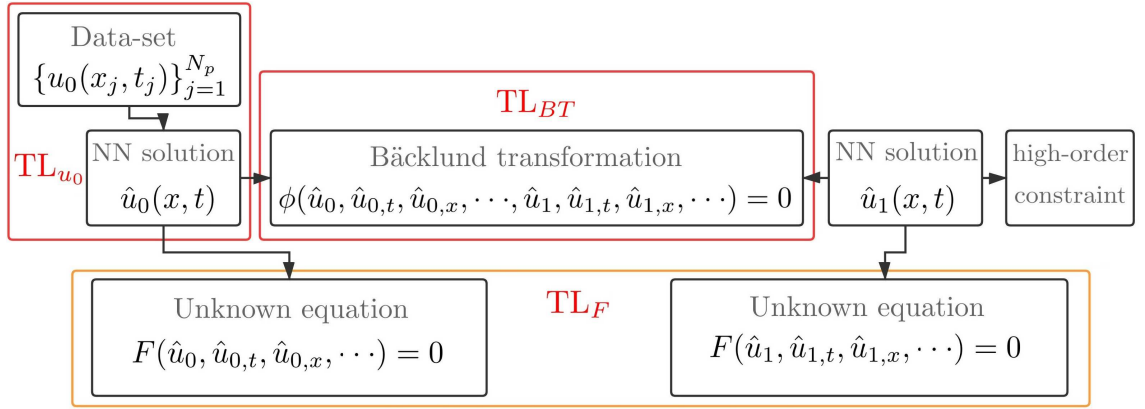}}}
\end{center}
\par
\vspace{0.05in}
\caption{\small The BT-enhanced PDE discovery scheme for the implicit BTs.}
\label{fig8-DNN3}
\end{figure}

\v {\it Case 2.}---Fig~\ref{fig8-DNN3} displays the case that the BT can be written as an implicit form. Since the
BT is an implicit form, thus we can not get the data-set of $u_1(x,t)$. We use the training loss ${\rm TL}_{BT}$ to obtain the approximated neural network solution $\hat{u}_1(x,t)$. And the approximated neural network solution $\hat{u}_0(x,t)$ can still be trained by the data-set $\{u_0(x_j,t_j)\}_{j=1}^{N_p}$ through ${\rm TL}_{u_0}$. The unknown equation will be learned by ${\rm TL}_{F}$. The correct equation will be
found more precise by the above two schemes. These models are trained by the efficient Adam optimization algorithm and L-BFGS optimization algorithm.

 The loss function of this scheme is
\bee
  {\rm TL}={\rm TL}_{u_0}+{\rm TL}_{BT}+{\rm TL}_{F},
\ene
where
\bee
 &&{\rm TL}_{u_0}=\displaystyle\frac{1}{N_p}\sum_{j=1}^{N_p}\left|u_0(x_j,t_j)-\hat{u}_0(x_j,t_j)\right|^2,\v\\
 &&{\rm TL}_{BT}=\displaystyle\frac{1}{N_p}\sum_{j=1}^{N_p}\left|\phi(\hat{u}_0(x_j,t_j),\hat{u}_1(x_j,t_j))\right|^2,\v\\
 &&{\rm TL}_{F}=\displaystyle\frac{1}{N_p}\sum_{j=1}^{N_p}\left(
 \left|F(\hat{u}_0(x_j,t_j))\right|^2+ \left|F(\hat{u}_1(x_j,t_j))\right|^2\right).
\ene
The exact equation will be found more precise by the above scheme. If one wants to obtain the higher accuracies of unknown equations, an additional neural network solution $\hat{u}_2$ should be added. The losses ${\rm TL}_{BT}$ and  ${\rm TL}_{F}$ in the TL are replaced  by
\bee
\begin{array}{rl}
 {\rm TL}_{BT^2}=\displaystyle\frac{1}{N_p}\sum_{j=1}^{N_p}\left(
 \left|\phi(\hat{u}_0(x_j,t_j),\hat{u}_1(x_j,t_j))\right|^2+\left|\phi(\hat{u}_1(x_j,t_j),\hat{u}_2(x_j,t_j))\right|^2\right)
\end{array}
\ene
and
\bee\begin{array}{rl}
{\rm TL}_{F}=&
\displaystyle\frac{1}{N_p}\sum_{j=1}^{N_p}\left(
 \left|F(\hat{u}_0(x_j,t_j))\right|^2+ \left|F(\hat{u}_1(x_j,t_j))\right|^2+ \left|F(\hat{u}_2(x_j,t_j))\right|^2\right).
\end{array}
\ene

In the two above schemes, if we only use $\hat{u}_0$ and $\hat{u}_1$ to train the neural network, then we call it the 1-fold BT-enhanced (BTE) scheme. And if we use $\hat{u}_0$, $\hat{u}_1$ and $\hat{u}_2$ in the training process, we call it the 2-fold BTE scheme. All schemes are trained by the efficient Adam  and L-BFGS optimization algorithms.

In what follows, we will display two examples to show the validity of our schemes. The first example is to learn the mKdV equation based on the explicit BT (alias Darboux transform (DT)) of the mKdV equation given by Eq.~(\ref{BT1}). The second example is to study the s-G equation via the implicit aBT (\ref{sG-back}).

 \begin{table}[!t]
	\centering
    \setlength{\tabcolsep}{5pt}% column separation
    \renewcommand{\arraystretch}{1.4}%row space
	\caption{Comparisons of PINNs and BT-enhanced scheme for the data-driven  mKdV equation discovery.   \vspace{0.35in}}
	\begin{tabular}{lcccccccccc} \hline\hline
	Case & $a$ & error of $a$ & $b$ & error of $b$ & $c$ & error of $c$ & $d$ & error of $d$ \\  \hline
  Exact & 6 & 0 & 1 & 0 & 0 & 0& 0 & 0 \\
  A (PINNs) & 5.63392 & 0.366 & 0.90826 & 0.0917 & 0 & 0 & 0 & 0 \\
  A (PINNs $2\%$) & 5.60595 & 0.394 & 0.90041 & 0.0996 & 0 & 0 & 0 & 0 \\
  A (PINNs $5\%$) & 5.56262 & 0.437 & 0.89004 & 0.110 & 0 & 0 & 0 & 0 \\
  A (PINNs $10\%$) & 5.58002 & 0.420 & 0.89078 & 0.109 & 0 & 0 & 0 & 0 \\
  A (PINNs $20\%$) & 3.64644 & 2.35356 & 0.45905 & 0.541 & 0 & 0 & 0 & 0 \\
  B (PINNs) & 5.58765 & 0.412 & 0.90205 & 0.0979 & -0.02825 & 0.0282 & -0.01493 & 0.0149 \\
  B (PINNs $2\%$) & 5.80756 & 0.192 & 0.95514 & 0.0449 & -0.02467 & 0.0247 & -0.02322 & 0.0232 \\
  B (PINNs $5\%$) & 5.74998 & 0.250 & 0.94189 & 0.0581 & -0.01166 & 0.0117 & -0.01847 & 0.0185 \\
  B (PINNs $10\%$) & 5.63770 & 0.362 & 0.91263 & 0.0874 & -0.00343 & 0.00343  & -0.01174 & 0.0117 \\
  B (PINNs $20\%$) & 4.86428 & 1.14 & 0.73158 & 0.268 & -0.02057 & 0.0206 & -0.00608 & 0.00608 \\
  A (BTE) & 6.00973 & 9.73$\times10^{-3}$ & 1.00123 & 1.23$\times10^{-3}$ & 0 & 0 & 0 & 0 \\
  A (BTE $2\%$) & 5.97970 & 2.03$\times10^{-2}$ & 0.99380 & 6.20$\times10^{-3}$ & 0 & 0 & 0 & 0 \\
  A (BTE $5\%$) & 5.98296 & 1.70$\times10^{-2}$ & 0.99231 & 7.69$\times10^{-3}$ & 0 & 0 & 0 & 0 \\
  A (BTE $10\%$) & 5.99462 & 5.38$\times10^{-3}$ & 0.99104 & 8.96$\times10^{-3}$ & 0 & 0 & 0 & 0 \\
  A (BTE $20\%$) & 5.97702 & 2.30$\times10^{-2}$ & 0.97901 & 2.10$\times10^{-2}$ & 0 & 0 & 0 & 0 \\
  B (BTE) & 5.99109 & 8.91$\times10^{-3}$ & 0.99804 & 1.96$\times10^{-3}$ & 0.00086 & 8.57$\times10^{-4}$ & 0.00110 & 1.10$\times10^{-3}$ \\
  B (BTE $2\%$) & 5.99943 & 5.72$\times10^{-4}$ & 0.99818 & 1.82$\times10^{-4}$ & 0.00013 & 1.35$\times10^{-4}$ & 0.00028 & 2.80$\times10^{-4}$ \\
  B (BTE $5\%$) & 5.98173 & 1.83$\times10^{-2}$ & 0.99214 & 7.86$\times10^{-4}$ & 0.00024 & 2.38$\times10^{-4}$ & 0.00016 & 1.62$\times10^{-4}$ \\
  B (BTE $10\%$) & 5.98347 & 1.65$\times10^{-2}$ & 0.98876 & 1.12$\times10^{-2}$ & -0.00063 & 6.26$\times10^{-4}$ & -0.00142 & 1.42$\times10^{-3}$ \\
  B (BTE $20\%$) & 5.98058 & 1.94$\times10^{-2}$ & 0.97879 & 2.12$\times10^{-2}$ & -0.00178 & 1.78$\times10^{-3}$ & -0.00305 & 3.05$\times10^{-3}$ \\
    \hline\hline
	\end{tabular}
	\label{mKdV-table2}
\end{table}

\begin{figure}[!t]
\hspace{2.5in}
\begin{center}
 \hspace{0.2in}{\scalebox{0.7}[0.65]{\includegraphics{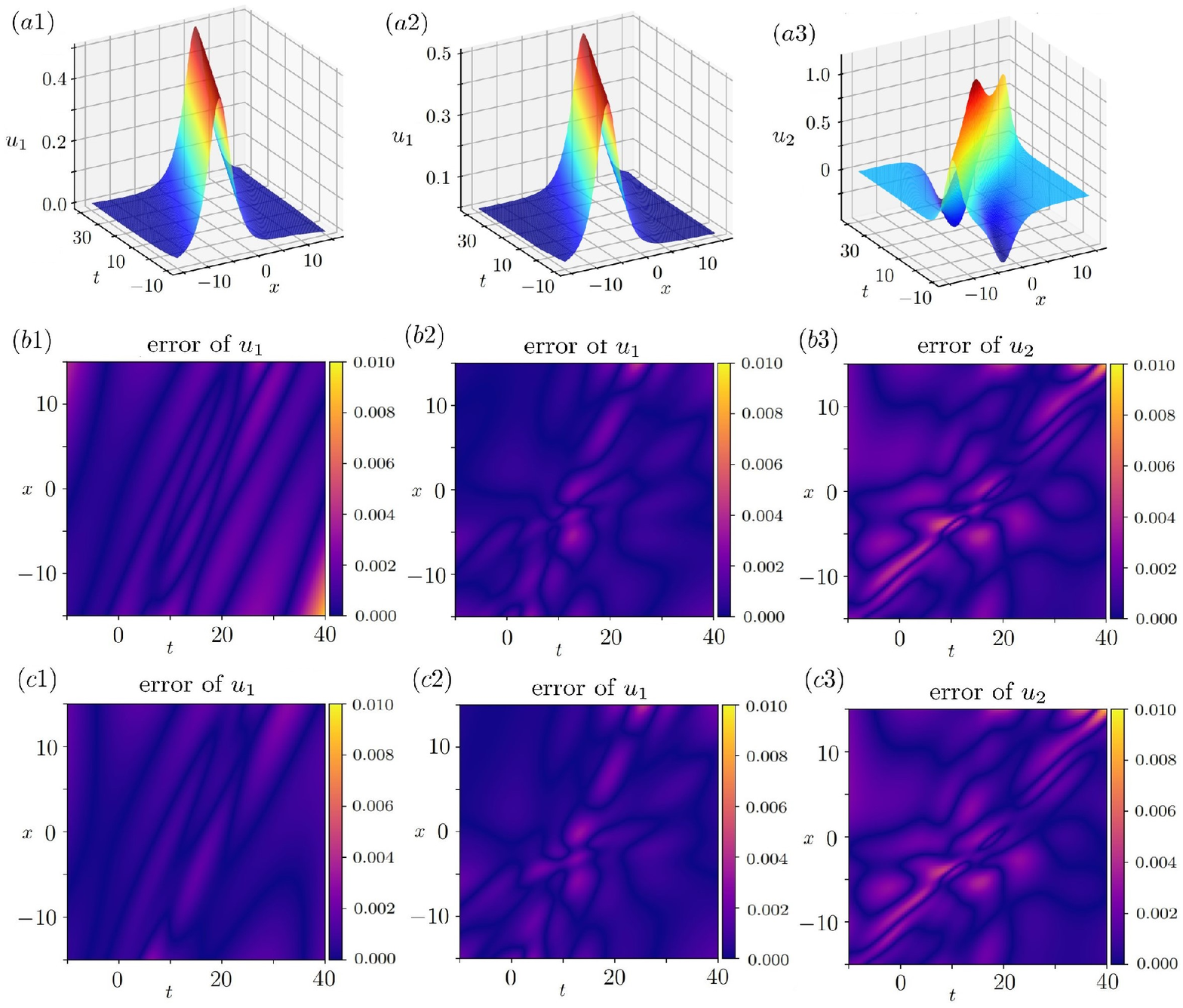}}}
\end{center}
\par
\vspace{-0.05in}
\caption{\small Focusing mKdV equation. (a1) trained one-soliton solution via PINNs and (a2, a3) trained one- and two-soliton
 solutions via the BT-enhanced PDE scheme.  (b1-c3) absolute errors: (b1) PINNs learning Case A with a 5$\%$ noise, (b2-b3)
 BT-enhanced PDE scheme learning Case A with a 5$\%$ noise, (c1) PINNs learning Case B with a 5$\%$ noise, (c2-c3) BT-enhanced PDE scheme learning Case B with a 5$\%$ noise. The relative $\mathbb{L}^2-$norm errors of $u_1$ and $u_2$ are: (b1) $6.61\times 10^{-3}$, (b2) $3.71\times 10^{-3}$, (b3) $4.61\times 10^{-3}$, (c1) $5.81\times 10^{-2}$, (c2) $3.80\times 10^{-3}$, and (c3) $4.41\times 10^{-3}$, respectively. The training times are (b1) 132.14s, (b2-b3) 302.67s, (c1) 135.95s, and (c2-c3) 285.60s, respectively.}
\label{fig9-mKdVsolution}
\end{figure}

\subsection{Data-driven discovery of  mKdV equation via the explicit BT/DT}

 Here we would like to use the known solution $u_1$, and the explicit BT/DT (\ref{BT1}) generating the new solution $u_2$ from $u_1$ to discover the unknown mKdV equation (\ref{mKdV3}) with some perturbation terms via the BT-enhanced scheme.

The focusing mKdV equation (\ref{mKdV}) possesses the BT/DT~\cite{gu2005}
\bee\label{BT1}
u_{n+1}=u_n+\frac{4\lambda\psi}{1+\psi^2},\quad \psi=\frac{\Phi_{22}(x,t,\lambda)+\mu\Phi_{21}(x,t,\lambda)}{\Phi_{12}(x,t,\lambda)+\mu\Phi_{11}(x,t,\lambda)}
\ene
with a free parameter $\mu$, where
\bee \no
\Phi=\begin{pmatrix} \Phi_{11}(x,t,\lambda)&\Phi_{12}(x,t,\lambda) \v\\
\Phi_{21}(x,t,\lambda)&\Phi_{22}(x,t,\lambda) \end{pmatrix}
\ene
is the basic solution of the Lax pair of the mKdV equation (\ref{mKdV})
\bee\label{mKdV-Lax}
%\begin{cases}
\Phi_x=\begin{pmatrix} \lambda & u \v\\ -u & -\lambda \end{pmatrix}\Phi, \quad
\Phi_t=\begin{pmatrix} -2\lambda(2\lambda^2+u^2) & -2\lambda(2u \lambda+u_x)-2u^3-u_{xx} \v\\
2\lambda(2u \lambda-u_x)+2u^3+u_{xx} & 2\lambda(2\lambda^2+u^2) \end{pmatrix}\Phi
%\end{cases}
\ene
with $\lambda\in\mathbb{C}$ being a spectral parameter, and $u$  an initial solution of the mKdV equation (\ref{mKdV}).

By using the above DT (\ref{BT1}) with $u_0=0,\, \mu=e^{2\alpha_1}$ and $\lambda=\lambda_1$, one can obtain the one-soliton solution of the mKdV equation (\ref{mKdV})
\bee\label{1soli}
u_1(x,t)=2\lambda_1{\rm sech}(2\lambda_1x-8\lambda_1^3t+2\alpha_1)
\ene
Further, one can use the DT (\ref{BT1}) with $u_1$ given by Eq.~(\ref{1soli}),  $\mu=e^{2\alpha_2}$ and $\lambda=\lambda_2$,  to find the 2-soliton solution of the mKdV equation (\ref{mKdV}) in the form
\bee\label{2soli}
u_2(x,t)=\frac{2(\lambda_2^2-\lambda_1^2)(\lambda_2{\rm cosh}(v_1)-\lambda_1{\rm cosh}(v_2))}
{(\lambda_1^2+\lambda_2^2){\rm cosh}(v_1){\rm cosh}(v_2)-2\lambda_1\lambda_2(1+{\rm sinh}(v_1){\rm sinh}(v_2))},
\ene
where $v_j=2\lambda_jx-8\lambda_j^3t+2\alpha_j,\,(j=1,2)$, and $\lambda_2>\lambda_1>0$. In this example,
we take $\lambda_1=\frac{1}{4}$, $\lambda_2=\frac{1}{3}$, $\alpha_1=1$, and $\alpha_2=2$. According to the  homogeneity principle of the mKdV equation, we add two disturbance terms $u^4$, $uu_{xx}$ to the mKdV equation to generate the generalized form
\bee\label{mKdV3}
 u_t+au^2u_x+bu_{xxx}+cu^4+duu_{xx}=0,
\ene
which is used to examine the robustness of the scheme, where $a,\, b,\,c$, and $d$  are real-valued parameters to be determined.

In what follows, we would like to use the above-mentioned BT-enhanced PDE discovery scheme given by Fig.~\ref{fig7-DNN2} to discover these parameters $a,\,b,\,c,\,d$ of the mKdV equation (\ref{mKdV3}) in two cases by considering the system
\bee\label{mkdv-dt}
\left\{\begin{array}{l}
 u_t+au^2u_x+bu_{xxx}+cu^4+duu_{xx}=0, \vspace{0.05in} \\
 u_2-u_1-\dfrac{4\lambda\psi}{1+\psi^2}=0.
   %\vspace{0.05in}\\
  %u(x,0)=u_0,\quad u(-L, x)=u(L,x),\vspace{0.05in}\\
  %v(x,0)=v_0(x),\quad v(-L, x)=v(L,x)
\end{array}\right.
\ene
 The training data-set is generated by the soliton solution (\ref{1soli}) of the focusing mKdV equation (\ref{mKdV}). To compare our scheme with the known PINNs scheme~\cite{raiss19}, which is trained by considering Eq.~(\ref{mKdV3}) with the initial data generating from $u_1$.

\begin{figure}[!t]
\hspace{2.5in}
\begin{center}
 {\scalebox{0.45}[0.45]{\includegraphics{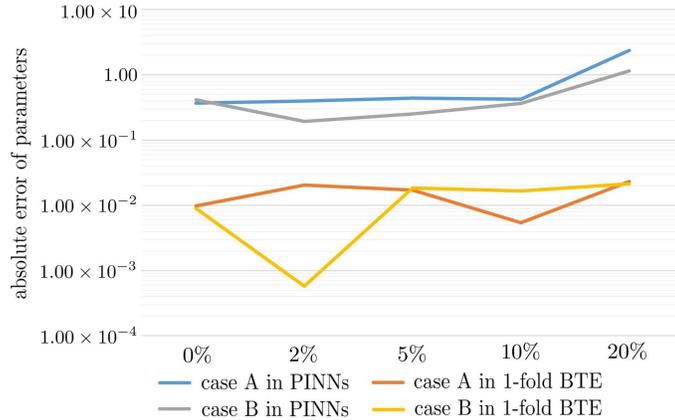}}}
\end{center}
\par
\vspace{-0.05in}
\caption{\small Error comparisons between the BTE and PINNs schemes for the mKdV equation. The vertical axis is the maximum of the absolute errors of all parameters.}
\label{fig10-mKdVcompare}
\end{figure}

\v {\it Case A.}---We fix $c=d=0$, and learn two unknown parameters $a$ and $b$. We take the initial values of $a,\,b$ as $a=b=1$. The training data-set is generated by not only the one-soliton solution (\ref{1soli}) but also the two-soliton solution (\ref{2soli}) of
the mKdV equation. The training data-set is sampled in
  the region $(x,t)\in[-15,15]\times[-10,40]$, and 10,000 sampling points are used in the training processes. For the  BT-enhanced PDE discovery scheme given by Fig.~\ref{fig7-DNN2}, two solitons $u_1$ and $u_2$ are approximated by using a 6-layer neural network with 40 neurons per layer, and 5,000 steps Adam and 5,000 steps L-BFGS optimizations are chosen. We use the usual PINNs with the training data arising from the one-soliton solution $u_1$, and our scheme with the same neural network to learn two parameters $a,\, b$, respectively, such that Case A of Table~\ref{mKdV-table2} displays the learning results and their errors under two senses of the training data without a noise and with the differential $2\%,\, 5\%,\, 10\%,\,20\%$ noises. Figs.~\ref{fig9-mKdVsolution} and \ref{fig10-mKdVcompare} exhibit the errors of training results of all conditions, which imply that our scheme is better than the PINNs scheme for the same training steps. That is to say, except for the one-soliton data arising from $u_1$, the used more data $u_2$ generated from the DT (\ref{BT1}) can make the errors of learning results smaller in the BT-enhanced PDE discovery scheme given by Fig.~\ref{fig7-DNN2}.

\v {\it Case B.}---We train all four unknown parameters $a,\,b,\, c,\, d$ in Eq.~(\ref{mKdV3}). We used the same BT-enhanced PDE discovery scheme given by Fig.~\ref{fig7-DNN2} as Case A and PINNs to study this case, respectively. For the two senses of the training data without a noise and with the differential $2\%,\, 5\%,\, 10\%,\,20\%$ noises, Case B in Table~\ref{mKdV-table2} displays the learning parameters about $a,\,b,\,c,\, d$, and their errors via our scheme and the known PINNs. The results imply that the BT-enhanced PDE discovery scheme given by Fig.~\ref{fig7-DNN2} is more effective (see Figs.~\ref{fig9-mKdVsolution} and \ref{fig10-mKdVcompare}).
 %Moreover, the errors are exhibited in Figs.~\ref{fig2-SGerror}(b3, b4) for the cases without a noise and with a $2\%$ noise, respectively.

\begin{table}[!t]
	\centering
    \setlength{\tabcolsep}{21pt}% column separation
    \renewcommand{\arraystretch}{1.4}%row space
	\caption{Comparisons of PINNs and BT-enhanced PDE scheme for the data-driven s-G equation discovery.   \vspace{0.1in}}
	\begin{tabular}{lcccccc} \hline\hline
	Case & $a$ & error of $a$ & $b$ & error of $b$ \\  \hline
  Exact & 1 & 0 & 0 & 0  \\
  PINNs (no noise) & 0.99991 & 9.18$\times10^{-5}$ & -0.00002 & 2.16$\times10^{-5}$ \\
  PINNs ($2\%$ noise) & 1.00051 & 5.13$\times10^{-4}$ & -0.00005 & 5.30$\times10^{-5}$  \\
  PINNs ($5\%$ noise) & 1.00128 & 1.28$\times10^{-3}$ & 0.00002 & 1.58$\times10^{-5}$  \\
  PINNs ($10\%$ noise) & 1.00239 & 2.39$\times10^{-3}$ & -0.00019 & 1.90$\times10^{-4}$ \\
  PINNs ($20\%$ noise) & 1.00466 & 4.66$\times10^{-3}$ & -0.00041 & 4.13$\times10^{-4}$  \\
  1-fold BTE (no noise)  & 0.99996 & 4.15$\times10^{-5}$ & 0.00001 & 1.30$\times10^{-5}$ \\
  1-fold BTE ($2\%$ noise) & 1.00020 & 1.98$\times10^{-4}$ & -0.00000 & 7.21$\times10^{-6}$  \\
  1-fold BTE ($5\%$ noise) & 1.00057 & 5.70$\times10^{-4}$ & -0.00003 & 2.97$\times10^{-5}$  \\
  1-fold BTE ($10\%$ noise) & 1.00108 & 1.07$\times10^{-3}$ & -0.00007 & 6.46$\times10^{-5}$ \\
  1-fold BTE ($20\%$ noise) & 1.00218 & 2.18$\times10^{-3}$ & -0.00010 & 9.67$\times10^{-5}$  \\
  2-fold BTE (no noise) & 0.99984 & 1.59$\times10^{-4}$ & -0.00004 & 4.05$\times10^{-5}$  \\
  2-fold BTE ($2\%$ noise) & 1.00003 & 3.42$\times10^{-5}$ & 0.00002 & 1.54$\times10^{-5}$  \\
  2-fold BTE ($5\%$ noise) & 1.00036 & 3.58$\times10^{-4}$ & -0.00004 & 3.72$\times10^{-5}$  \\
  2-fold BTE ($10\%$ noise) & 1.00081 & 8.08$\times10^{-4}$ & -0.00001 & 1.10$\times10^{-5}$  \\
  2-fold BTE ($20\%$ noise) & 1.00146 & 1.46$\times10^{-3}$ & -0.00011 & 1.12$\times10^{-4}$ \\
    \hline\hline
	\end{tabular}
	\label{sG-table3}
\end{table}

\subsection{Data-driven discovery of the s-G equation via the implicit aBT}

 In what follows, we would like to use the known breather solution (\ref{sG-solu}) and implicit aBT (\ref{sG-back}) to discover the unknown s-G equation (\ref{sG1}) with some perturbation terms via the BT-enhanced PDE discovery scheme.

We know that the s-G equation (\ref{sG}) has the aBT (\ref{sG-back}), which can be used to generate its new solution $u'(x,t)$.
In what follows, we will utilize this aBT to find the high-precision s-G equation. Meanwhile, we will compare the results between the
BT-enhanced PDE discovery scheme given by Fig.~\ref{fig8-DNN3} and known PINNs scheme~\cite{raiss19}. We consider the aBT of the generalized s-G equation
\bee\label{sG1}
 u_{xt}=a\sin u+b\cos u,
\ene
where $a,\, b$ are parameters to be determined. As $a=1,\, b=0$, Eq.~(\ref{sG1}) reduces to the known aBT of the s-G equation\cite{sG}.

In what follows, we would like to use the above-mentioned BT-enhanced PDE discovery scheme given by Fig.~\ref{fig8-DNN3} to discover these parameters $a,\,b$ of the generalized s-G equation (\ref{sG1}) in two cases by considering the system
\bee\label{mkdv-dt}
\left\{\begin{array}{l}
 u_{xt}-a\sin u-b\cos u=0, \vspace{0.05in} \\
 u'_x=u_x-2\beta\sin\left(\dfrac{u+u'}{2}\right), \v \\
 u'_t=-u_t+\dfrac{2}{\beta}\sin\left(\dfrac{u-u'}{2}\right).
   %\vspace{0.05in}\\
  %u(x,0)=u_0,\quad u(-L, x)=u(L,x),\vspace{0.05in}\\
  %v(x,0)=v_0(x),\quad v(-L, x)=v(L,x)
\end{array}\right.
\ene
 The training data-set is generated by the breather solution (\ref{sG-solu}) of the s-G equation (\ref{sG}). To compare our scheme with the known PINNs scheme, which is trained by considering Eq.~(\ref{sG1}) with the initial data generating from the breather solution (\ref{sG-solu}).

\begin{figure}[!t]
\hspace{2.5in}
\begin{center}
 \hspace{0.2in}{\scalebox{0.6}[0.6]{\includegraphics{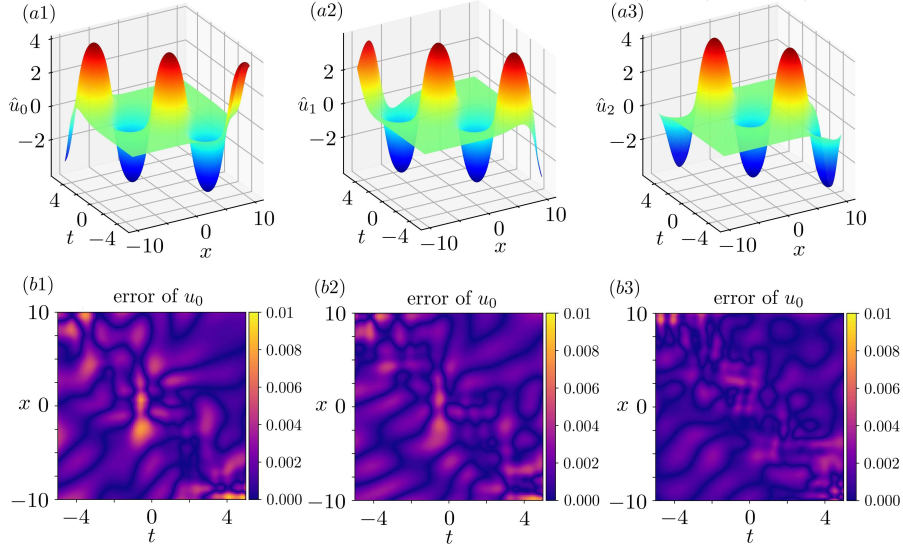}}}
\end{center}
\par
\vspace{-0.05in}
\caption{\small(a1-a3) Three neural network solutions of s-G equation via the 2-fold BTE scheme. (b1-b3) absolute errors
of $u1$ with a 2$\%$ noise: (b1) PINNs scheme, (b2) 1-fold BTE  scheme, (b3) 2-fold BTE  scheme. The relative $\mathbb{L}^2-$norm errors of $u_1$ and training times are (b1) $1.51\times 10^{-3}$, 123.20s, (b2) $1.33\times 10^{-3}$, 343.72s, (b3)$1.01\times 10^{-3}$, 505.11s, respectively. }
\label{fig11-SGsolution}
\end{figure}

\begin{figure}[!t]
\hspace{2.5in}
\begin{center}
 \hspace{0.2in} {\scalebox{0.45}[0.45]{\includegraphics{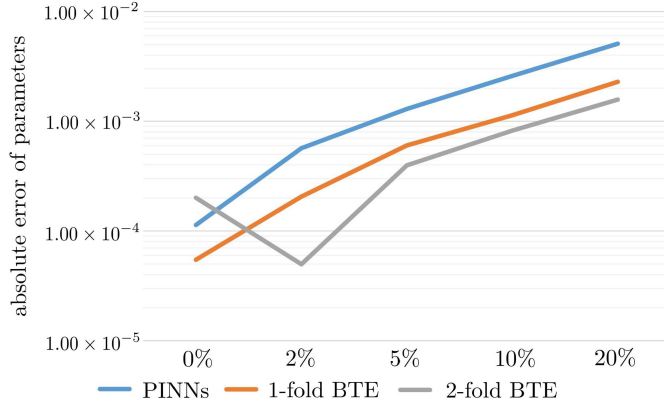}}}
\end{center}
\par
\vspace{-0.05in}
\caption{\small  Error comparisons between the BTE scheme and PINNs for the sine-Gordon equation via the implicit aBT. The vertical axis is the maximum of the absolute errors of all parameters.}
\label{fig12-SGcompare}
\end{figure}

The training data are generated by the breather solution (\ref{sG-solu}). The 10,000 sampling points are selected in $\{(x,t)|(x,t)\in[-10,10]\times[-5,5]\}$. The 5,000 steps Adam and 5,000 steps L-BFGS optimizations will be used in each case.
the trained solutions $\hat{u}_0$, $\hat{u}_1$ and $\hat{u}_2$ are represented by using one 6-layer neural network with 40 neurons per layer. $\hat{u}_0$ will approach to the right solution (\ref{sG-solu}), and $\hat{u}_1$, $\hat{u}_2$ just are the neural network solutions. Fig.~\ref{fig11-SGsolution} displays the breather-like profiles of $\hat{u}_0$, $\hat{u}_1$ and $\hat{u}_2$. Fig.~\ref{fig12-SGcompare} exhibits the training results of all conditions, where the vertical axis represents the maximal errors of $b$ and $c$, and the horizontal axis denotes the different error of training data. As a result, we find that our scheme is
almost always better than the PINNs scheme for the same training steps.


\section{Conclusions and discussions}

 In conclusion, we have established the deep neural network learning algorithms to discover the BTs and soliton evolution equations. Three types of BTs are used to investigate the availability of our schemes, such as the aBT of the sine-Gordon equation, complex/real Miura transform between the defocusing/focusing mKdV equation and KdV equation, as well as the DT of the focusing mKdV equation. Moreover, we also compare our scheme with the known PINNs such that our scheme is more effective in the study of data-driven discoveries of soliton equations via the BTs. The idea can also be extended to other types of BTs and nonlinear evolution PDEs containing some physically interesting soliton evolution equations.

%\v \noindent {\bf Declaration of Competing Interest}

%\v The authors declare that they have no known competing financial interests or personal relationships that could have appeared to influence the work reported in this paper.

\v \noindent {\bf Acknowledgements} \v

%The authors would like to thank the referees for the valuable suggestions and comments to improve the manuscript.
This work was supported by the National Natural Science Foundation of China (Nos. 11925108 and 11731014).

\v\noindent {\bf Data availability}\v

All data included in this study are available upon request by contact with the corresponding author.

%\v\noindent {\bf Declaration of competing interest} \v

\v\noindent {\bf Declarations}

\v\noindent {\bf Conflict of interest }\v

The authors declare no conflict of interest.

\end{document}